\documentclass[journal]{IEEEtran}

\usepackage{xcolor,soul,framed} 

\colorlet{shadecolor}{yellow}
\usepackage[pdftex]{graphicx}
\graphicspath{{../pdf/}{../jpeg/}}
\DeclareGraphicsExtensions{.pdf,.jpeg,.png}

\usepackage[font=small]{caption}
\usepackage[cmex10]{amsmath}
\usepackage{array}
\usepackage{mdwmath}
\usepackage{multirow}
\usepackage{booktabs, siunitx}
\usepackage{mdwtab}
\usepackage{eqparbox}
\usepackage{url}
\usepackage{amsfonts}

\usepackage{amssymb}
\usepackage{pifont}
\newcommand{\cmark}{\ding{51}}%
\newcommand{\xmark}{\ding{55}}%

\usepackage{ulem}  
\usepackage{algorithm}
\usepackage{algpseudocode}
\usepackage{subcaption}
\usepackage{longtable}
\usepackage{multicol} 
\usepackage{xcolor}
\usepackage{tablefootnote}
\usepackage{todonotes}
\usepackage{float}
\usepackage{tabularx} 
\usepackage{hyperref}
\usepackage{dirtytalk}
\usepackage{morewrites}
\begin{document}
    \title{GRAFNet: Multiscale Retinal Processing via Guided Cortical Attention Feedback for Enhancing Medical Image Polyp Segmentation}
    
  \author{Abdul Joseph Fofanah, ~\IEEEmembership{Member,~IEEE,}
        Lian Wen,~\IEEEmembership{Member,~IEEE,}
        Alpha Alimamy Kamara,
        Zhongyi Zhang,
        David Chen,~\IEEEmembership{Member,~IEEE,}
        Albert Patrick Sankoh,~\IEEEmembership{Member,~IEEE}

 \thanks{This work was supported in part by Griffith University under Grant 58455.}
  \thanks{Abdul Joseph Fofanah, Lian Wen, David Chen, and Zhongyi Zhang are with the School of Information and Communication Technology, Griffith University, Brisbane, 4111, Australia. (e-mail:abdul.fofanah@griffithuni.edu.au, orcid=0000-0001-8742-9325;  email: l.wen@griffith.edu.au, orcid=0000-0002-2840-6884); email: d.chen@griffith.edu.au, orcid=0000-0001-8690-7196); email: zhongyi.zhang@griffithuni.edu.au, orcid=0000-0002-0851-9953)}

\thanks{Alpha Alimamy Kamara is with the School of Computer Science and Engineering, Central South University, Changsha 410083, China (email: kamara@csu.edu.cn, orcid=0009-0008-9252-5950)}

\thanks{Albert Patrick Sankoh is with the School of Computer Science and Engineering, Northeastern University, Boston, United State of America (email: sankoh.a@northeastern.edu, orcid=0009-0001-2618-0447)}

\thanks{\textit{Corresponding Author:}  abdul.fofanah@griffithuni.edu.au}
}

\markboth{IEEE Transactions on Circuits and Systems for Video Technology, August~2025}{Abdul Joseph \MakeLowercase{\textit{Fofanah et al.}}: FACBNet: Multiscale Retinal Processing via Guided Cortical Attention Feedback for Enhancing Medical Image Polyp Segmentation}

\maketitle

\begin{abstract}
Accurate polyp segmentation in colonoscopy is essential for cancer prevention but remains challenging due to: (1) high morphological variability (from flat to protruding lesions), (2) strong visual similarity to normal structures such as folds and vessels, and (3) the need for robust multi-scale detection. Existing deep learning approaches suffer from unidirectional processing, weak multi-scale fusion, and the absence of anatomical constraints, often leading to false positives (over-segmentation of normal structures) and false negatives (missed subtle flat lesions). We propose \textbf{GRAFNet}, a biologically inspired architecture that emulates the hierarchical organization of the human visual system. GRAFNet integrates three key modules: (1) a \textit{Guided Asymmetric Attention Module} (GAAM) that mimics orientation-tuned cortical neurons to emphasise polyp boundaries, (2) a \textit{MultiScale Retinal Module} (MSRM) that replicates retinal ganglion cell pathways for parallel multi-feature analysis, and (3) a \textit{Guided Cortical Attention Feedback Module} (GCAFM) that applies predictive coding for iterative refinement. These are unified in a \textit{Polyp Encoder-Decoder Module} (PEDM) that enforces spatial–semantic consistency via resolution-adaptive feedback. Extensive experiments on five public benchmarks (Kvasir-SEG, CVC-300, CVC-ColonDB, CVC-Clinic, and PolypGen) demonstrate consistent state-of-the-art performance, with 3–8\% Dice improvements and 10–20\% higher generalisation over leading methods, while offering interpretable decision pathways. This work establishes a paradigm in which neural computation principles bridge the gap between AI accuracy and clinically trustworthy reasoning. Code is available at: \url{https://github.com/afofanah/GRAFNet}.
\end{abstract}


\IEEEpeerreviewmaketitle

\section{Introduction}
\IEEEPARstart{M}{edical} image analysis has become a cornerstone of modern healthcare, with automated detection and segmentation systems showing strong potential to improve diagnostic accuracy and clinical workflow efficiency~\cite{khalifa2024ai,tavanapong2022artificial}. Among its applications, polyp segmentation in colonoscopy is one of the most consequential yet elusive challenges in medical artificial intelligence (AI)~\cite{jha2021real}. Clinicians must detect subtle lesions hidden within the colon’s complex landscape of folds, fluids, and reflections~\cite{tavanapong2022artificial}. Current AI systems, though promising, often fail where human experts succeed—distinguishing meaningful patterns from noise~\cite{rajpurkar2022ai}. The consequences are significant: missed polyps lead to delayed cancer diagnoses~\cite{jayasekara2017risk}, while false alarms disrupt clinical workflow~\cite{mosch2024alarm}. There is thus an urgent need for systems that combine algorithmic precision with a clinician’s contextual understanding~\cite{intro15}.

The human visual system offers a powerful design principle. It processes scenes through parallel pathways—some analysing fine details like a microscope~\cite{intro14}, others capturing global contours like a wide-angle lens~\cite{intro16}. Crucially, these pathways are coupled by feedback loops that allow the brain to resolve ambiguities by “asking for clarification”~\cite{voitov2022cortical}. This bidirectional communication between local and global processing suggests that segmentation models should also combine bottom-up and top-down reasoning~\cite{peng2024fine}.

\begin{figure}[t]
    \centering
    \includegraphics[width=\columnwidth]{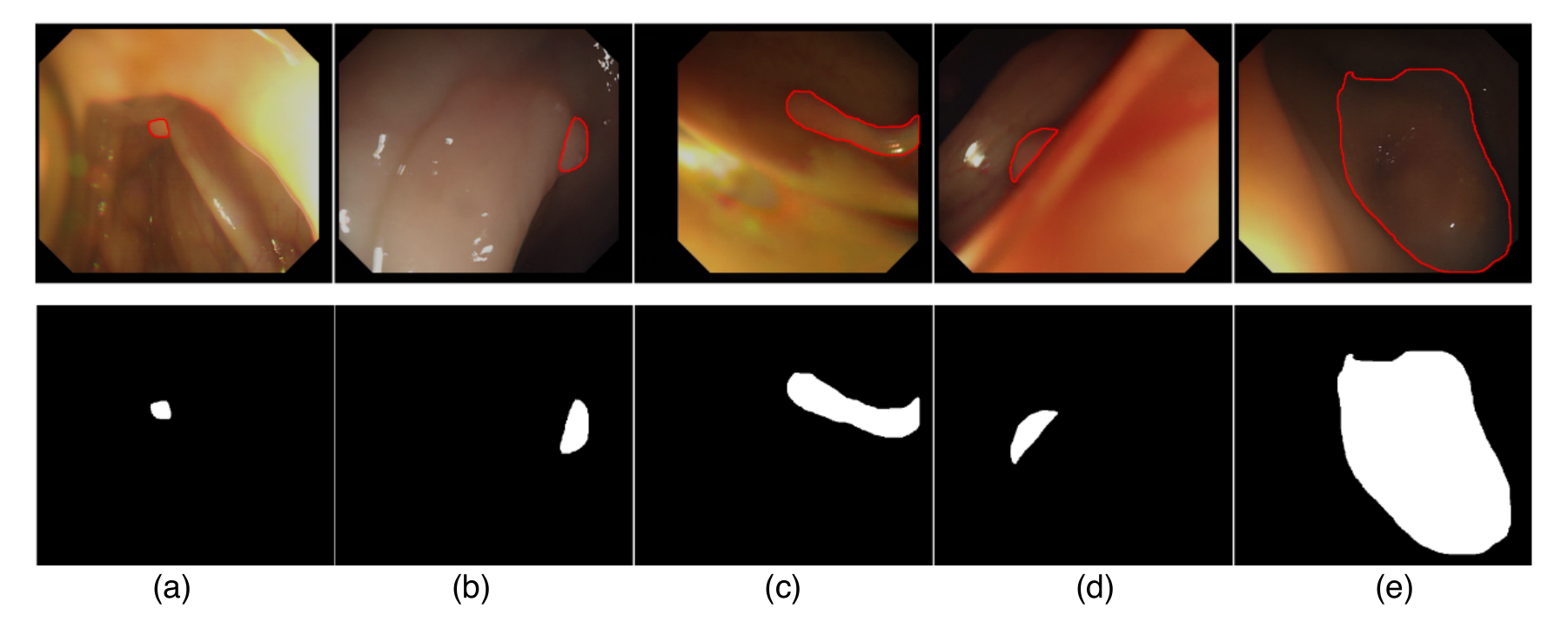}
    \caption{
    Diagnostic challenges in colonoscopy polyp segmentation.
    Top row: ground-truth masks overlaid in red. 
    Bottom row: binary ground-truth masks. 
    (a–b) Flat sessile polyps circled in red blending with mucosal folds. 
    (c–d) Polyps requiring complementary visual cues for accurate segmentation.}
    \label{fig:polyp_challenges}
\end{figure}

Most existing approaches resemble cameras with fixed focus: they either capture global structure while missing fine details~\cite{wang2025dcatnet}, or zoom in narrowly and lose anatomical context~\cite{jha2021comprehensive}. Typically, they process features in a single bottom-up direction, like an artist painting without ever stepping back to view the canvas~\cite{tudela2024complete}. This unidirectional flow leads to errors clinicians rarely make, such as confusing mucosal folds for polyps or missing flat lesions that blend with surrounding tissue~\cite{khan2025advances}. The missing element is visual cognition—the iterative refinement of interpretations~\cite{wang2025dcatnet}. As illustrated in Fig.~\ref{fig:polyp_challenges}, current systems often misclassify subtle sessile polyps (red) as folds or fail to detect them at all, because they cannot integrate micro-texture cues, 3D shape information, and anatomical context.

We propose \textbf{GRAFNet}, a biologically inspired architecture that mirrors three evolutionary-optimised mechanisms of the human visual system. The \textit{Guided Asymmetric Attention Module} (GAAM) replicates V1 cortical columns, using orientation-tuned units to enhance diagnostically relevant edges while suppressing noise~\cite{chen2025contour}. The \textit{Multiscale Retinal Module} (MSRM) models the retina’s parallel pathways, maintaining dedicated streams for fine textures (parvocellular), broad shapes (magnocellular), and colour contrast (koniocellular), similar to retinal ganglion cell processing~\cite{tam2021augmenting}. Finally, the \textit{Guided Cortical Attention Feedback Module} (GCAFM) implements predictive coding, dynamically refining low-level analysis with high-level anatomical priors. Together, these mechanisms form a closed-loop system that approximates the brain’s perception–refinement cycle.

The main contributions of this work are:
\begin{itemize}
    \item A Guided Asymmetric Attention Module (GAAM) that emulates orientation-tuned cortical neurons through steerable filters, selectively enhancing polyp boundaries while suppressing anatomical noise.
    \item A Multiscale Retinal Module (MSRM) that replicates retinal parallel processing with pathways (parvocellular, magnocellular, koniocellular, and ON–OFF) and lateral inhibition, enabling simultaneous texture, shape, and colour analysis while reducing redundancy.
    \item A Guided Cortical Attention Feedback Module (GCAFM) that applies predictive coding to resolve diagnostic ambiguities by reconciling low-level observations with high-level expectations.
    \item The unified GRAFNet architecture, which integrates GAAM, MSRM, and GCAFM within a Polyp Encoder–Decoder Module (PEDM) to maintain hierarchical consistency through resolution-adaptive feedback, preventing attention drift and delivering state-of-the-art segmentation performance. 
\end{itemize}

This paper is organised as follows: Section~\ref{sec:related_works} reviews related work; Section~\ref{sec:preliminary} introduces preliminaries and the problem formulation; Section~\ref{sec:method} details the methodology; Section~\ref{sec:experiments} presents experiments and results; and Section~\ref{sec:conclusion} concludes with findings and future directions.

\section{Related Works}
\label{sec:related_works}
This section reviews the evolution of polyp segmentation methods, from traditional techniques to modern deep learning and biologically inspired systems, and highlights the limitations that motivate our work.

\subsection{Traditional Polyp Segmentation Approaches}
Early methods relied on handcrafted features such as color and texture~\cite{shen2021cotr} or shape-based geometric priors~\cite{iakovidis2024effective}. Region-growing algorithms were also explored~\cite{ige2023convsegnet}. Although computationally efficient, these approaches lacked robustness and failed to generalise across diverse polyp morphologies and imaging conditions, limiting their clinical applicability~\cite{jha2021real}.

\subsection{Deep Learning Revolution}
The advent of deep learning transformed polyp segmentation. Fully Convolutional Networks (FCNs)~\cite{song2022fully} laid the foundation, with UNet~\cite{intro21} becoming a widely adopted baseline. Enhancements such as attention mechanisms~\cite{azad2024medical} and residual connections~\cite{wang2024osffnet} further improved performance. More recently, transformer-based architectures~\cite{ren2023ukssl,yue2025boundary} have captured long-range dependencies but at the cost of computational complexity. A common limitation across these models is their purely bottom-up, feedforward processing, which lacks the iterative, feedback-driven mechanisms of human vision.

\subsection{Biologically Inspired Vision Systems}
Neuroscience-inspired architectures provide new directions. Some replicate retinal parallel pathways for texture and shape analysis~\cite{bae2023study} or implement ON–OFF receptive fields~\cite{cai2023contour}, but they omit integrated feedback mechanisms~\cite{chimitt2024scattering}. Cortical-inspired systems such as predictive coding networks~\cite{millidge2022predictive} introduce top-down modulation but lack domain-specific constraints. Similarly, steerable filter-based approaches~\cite{shi2021unsharp} enhance contour detection but are not embedded within a complete visual hierarchy.

\subsection{Identified Gaps and Proposed Direction}
Three critical gaps remain: (1) the absence of architectures combining retinal parallel processing with cortical feedback loops, (2) the lack of anatomical constraints tailored to endoscopic imaging, and (3) the inability to maintain multi-scale consistency while preserving fine details. This work addresses these challenges through a novel architecture that integrates asymmetric attention blocks, multi-scale retinal modules, and guided cortical attention feedback for anatomically constrained predictive coding.

\section{Preliminary}
\label{sec:preliminary}

\begin{table}[!htb]
\centering
\caption{Core notations for the GRAFNet framework.}
\label{tab:notations}
\begin{tabular}{ll}
\toprule
\textbf{Notation} & \textbf{Description} \\
\midrule
$\mathbf{I} \in \mathbb{R}^{H \times W \times 3}$ & Input endoscopic image \\
$\mathbf{X} \in \mathbb{R}^{B \times C \times H \times W}$ & Feature map batch \\
$\mathbf{M}_{gt} \in \{0,1\}^{H \times W}$ & Ground truth segmentation mask \\
$B, C, H, W$ & Batch size, channels, height, width \\
$(x,y)$ & Spatial coordinate indices \\
$\mathcal{D}$ & Training dataset distribution \\
$\mathbf{A}_t \in \mathbb{R}^{B \times C \times H \times W}$ & Attention output at timestep $t$ \\
$\mathbf{G}, \mathbf{G}_{\text{cortical}}$ & Cortical guidance signals \\
$\mathbf{G}_{\text{feedback}} \in \mathbb{R}^{B \times C \times 1 \times 1}$ & Top-down gating signal \\
$\mathbf{G}_{\text{spatial}}, G_i$ & Spatial guidance, level-$i$ guidance \\
$\mathcal{A}_{\text{spatial}}, \mathcal{A}_{\text{channel}}$ & Spatial and channel attention \\
$\mathcal{A}_{\text{guided}}$ & Guided attention coordinates \\
$\text{Attn}_i$ & Attention at resolution level $i$ \\
$\hat{\mathbf{A}}(\mathbf{H})$ & Target attention from high-level features \\
$\mathcal{L}_{\text{BIO}}, \mathcal{L}_{\text{dice}}$ & Bio-inspired loss, Dice loss \\
$\mathcal{L}_{msrm}, \mathcal{L}_g$ & MSRM loss, guidance loss \\
$\mathcal{L}_{guid.}$ & Guidance component \\
$\boldsymbol{\theta}$ & Model parameters \\
$\alpha, \beta, \lambda, \eta$ & Loss weights \\
$\epsilon_{\text{feedback}}, \tau_{\text{alignment}}$ & Feedback/ alignment thresholds \\
\bottomrule
\end{tabular}
\end{table}

\subsection{Problem Formulation}
We frame polyp segmentation as a biologically inspired attention optimisation problem. Given input $\mathbf{I}$ and ground truth $\mathbf{M}_{gt}$, the task is to learn a segmentation function $\mathcal{S}$ addressing three challenges:

\textit{Challenge 1 (Scale Diversity):} Polyps vary from very small ($<100$ pixels) to large ($>1000$ pixels). Fixed receptive fields $\mathbf{W}_{\text{fixed}}$ are inadequate for multi-scale analysis.  

\textit{Challenge 2 (Static Attention):} Most attention mechanisms are static and bottom-up, lacking adaptive guidance from higher-level semantics.  

\textit{Challenge 3 (Biological Plausibility):} Existing architectures overlook the hierarchical, feedback-driven nature of vision.  

Accordingly, GRAFNet ($F_m(\mathbf{I})$) minimises a bio-inspired loss:
\begin{equation}
    \min_{\boldsymbol{\theta}} \; \mathbb{E}_{(\mathbf{I}, \mathbf{M}_{gt}) \sim \mathcal{D}} 
    \Big[ \mathcal{L}_{\text{BIO}}\big(\mathcal{S}_{\boldsymbol{\theta}}(\mathbf{I}), \mathbf{M}_{gt}\big) \Big],
\end{equation}
subject to constraints on feedback consistency, retinal pathway integration, and attention guidance alignment.

\subsection{Notations}
The framework integrates four biologically inspired modules:
\begin{itemize}
    \item \textit{Asymmetric Attention Block (AAB):} Directional convolutions ($\mathcal{D} = \{\text{h}, \text{v}, \text{d}_1, \text{d}_2\}$) with centre–surround features for orientation-selective processing.  
    \item \textit{Multiscale Retinal Module (MSRM):} Visual pathways (parvocellular $\mathcal{L}_P$, magnocellular $\mathcal{L}_M$, koniocellular $\mathcal{L}_K$, and ON–OFF $\mathcal{L}_O$) to handle shape, motion, and colour.  
    \item \textit{Guided Cortical Attention Feedback (GCAFM):} Combines high-level ($H$), low-level ($L$), and attention ($A$) features for top-down feedback.  
    \item \textit{Polyp Encoder–Decoder (PEDM):} Hierarchical encoder–decoder with iterative refinement and resolution-adaptive guidance.  
\end{itemize}
Standard operators (e.g., $\odot$, $\oplus$, $\parallel$) and hyperparameters are listed in Table~\ref{tab:notations}.

\section{Methodology}
\label{sec:method}
\begin{figure*}[th]
    \centering
    \includegraphics[width=1\linewidth]{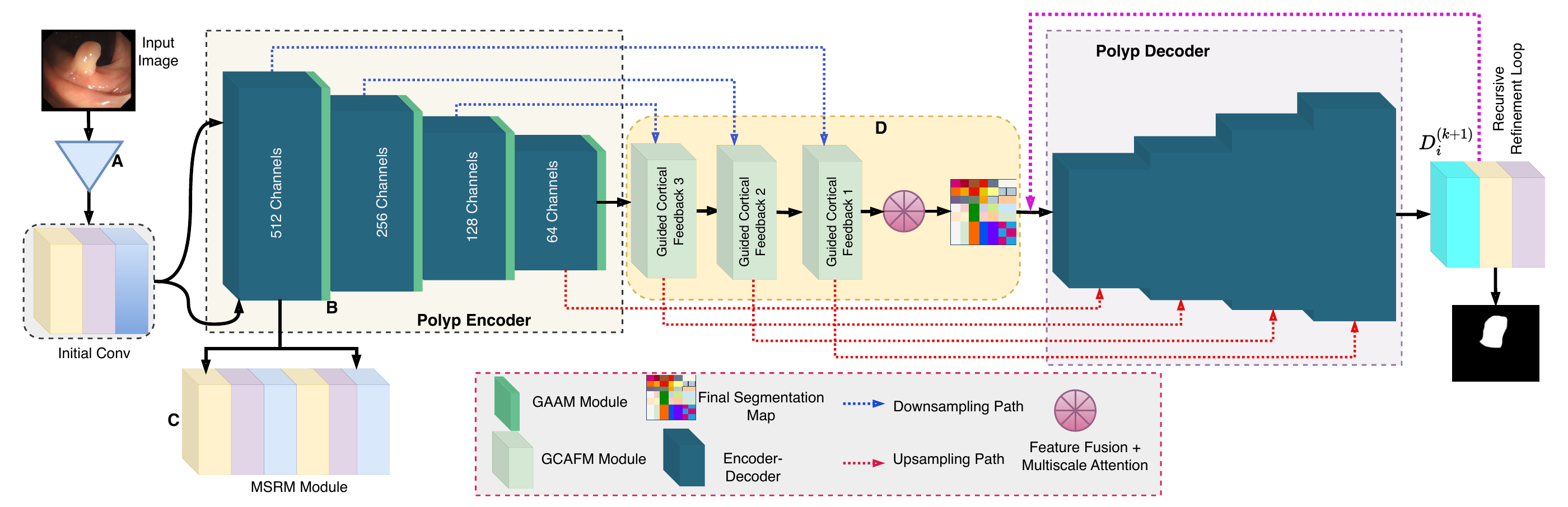}
    \caption{Overview of GRAFNet. The backbone (A) extracts visual features. GAAM (B) captures boundaries, MSRM (C) extracts multiscale features while reducing redundancy, and GCAFM (D) refines features using top-down feedback.}
    \label{fig:Overall_Architecture}
\end{figure*}

We describe \textit{GRAFNet} ($F_m(\mathbf{I})$), a biologically inspired framework designed to address scale diversity, static attention, and biological implausibility in polyp segmentation. As shown in Fig.~\ref{fig:Overall_Architecture}, the architecture integrates three modules that mimic human visual processing, from retinal encoding to cortical refinement.

\subsection{Guided Asymmetric Attention Module (GAAM)}
\begin{figure*}[th]
    \centering
    \includegraphics[width=1\linewidth]{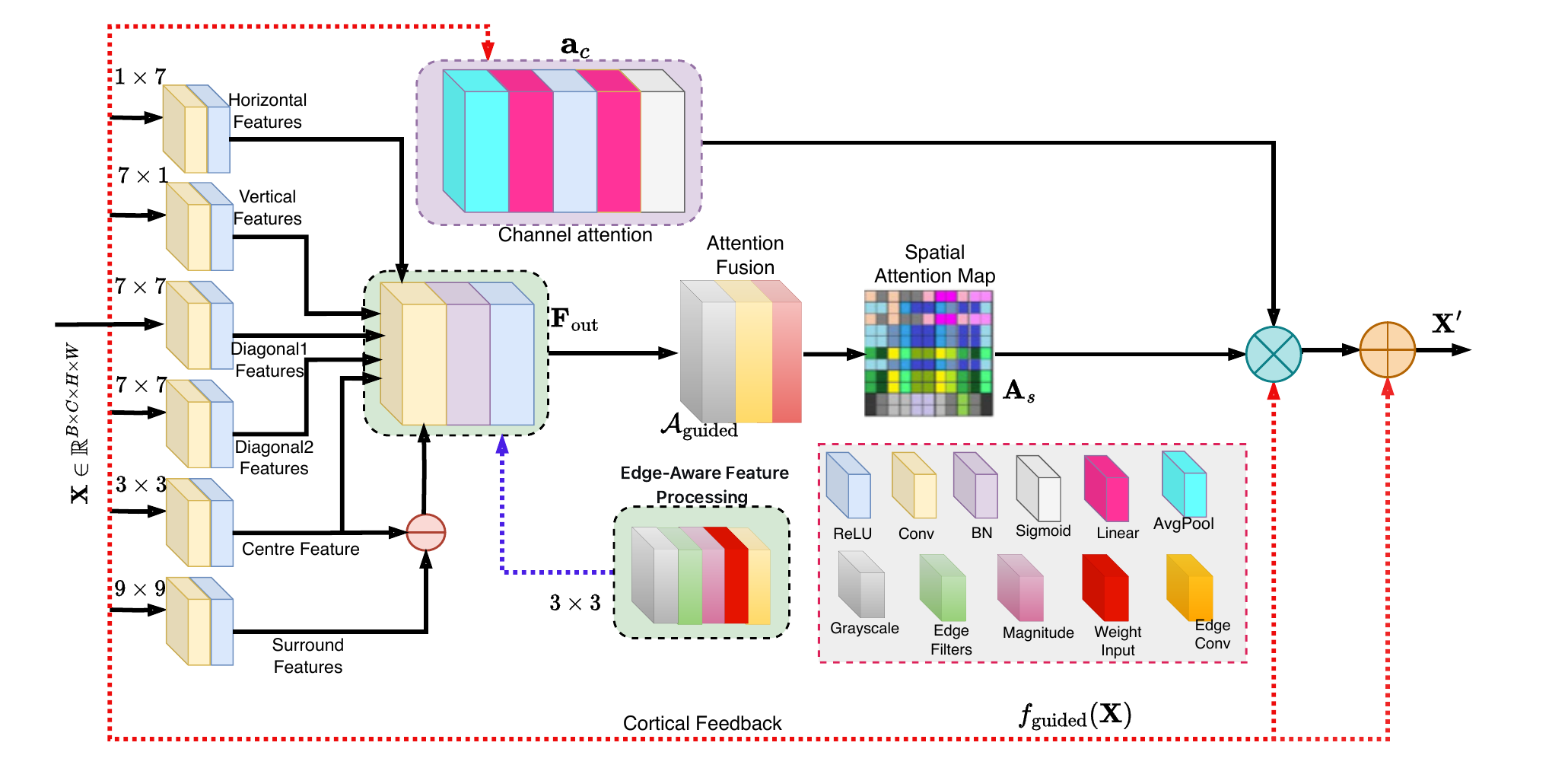}
    \caption{Guided asymmetric attention module.}
    \label{fig:gaam_module}
\end{figure*}

Static attention lacks adaptive, context-aware processing (\textit{Challenge 2}). GAAM explicitly enhances polyp boundaries and textures by mimicking orientation-sensitive neurons in the primary visual cortex (V1), as shown in Fig.~\ref{fig:gaam_module}.

The module processes $\mathbf{X}$ in parallel:  
(1) $\mathbf{F}_h = f_h(\mathbf{X})$ (1$\times$7 conv);  
(2) $\mathbf{F}_v = f_v(\mathbf{X})$ (7$\times$1 conv);  
(3–4) $\mathbf{F}_{d1} = f_{d1}(\mathbf{X}\odot \mathbf{M}_{d1})$, $\mathbf{F}_{d2} = f_{d2}(\mathbf{X}\odot \mathbf{M}_{d2})$ (masked 7$\times$7 convs);  
(5) $\mathbf{F}_{cs} = f_c(\mathbf{X}) - 0.6 f_s(\mathbf{X})$ (3$\times$3, 9$\times$9 kernels);  
(6) $\mathbf{F}_e = f_e(\mathbf{X}\odot \text{EdgeMagnitude}(\mathbf{X}))$ using Sobel and Laplacian filters.  

The features are fused into a spatial attention map:
\begin{equation}
\mathbf{A}_s = \sigma\left(f_{fusion}([\mathbf{F}_h,\mathbf{F}_v,\mathbf{F}_{d1},\mathbf{F}_{d2},\mathbf{F}_{cs},\mathbf{F}_e])\right),
\end{equation}
where $f_{fusion}$ is a series of $1\times1$ convolutions.  
Channel attention is computed as:
\begin{equation}
\mathbf{a}_c = \sigma\!\left(f_{c2}\big(\text{ReLU}(f_{c1}([\text{GAP}(\mathbf{X}),\text{GMP}(\mathbf{X})]))\big)\right),
\end{equation}
with GAP/GMP as global pooling, and $f_{c1},f_{c2}$ linear layers.  
The final output:
\begin{equation}
\mathbf{X}' = \mathbf{X} + (\mathbf{X}\odot \mathbf{A}_s \odot \mathbf{a}_c).
\end{equation}

Unlike static mechanisms, GAAM incorporates anatomical priors through dynamically weighted attention:
\begin{equation}
f_{\text{guided}}(\mathbf{X}) = \arg\max_{\theta} \sum_{d \in \mathcal{D}} w_d(\theta,\mathbf{C}) \cdot \text{Resp}_d(\mathbf{X},\theta),
\end{equation}
where $\mathcal{D}=\{\text{h},\text{v},\text{d}_1,\text{d}_2\}$ indexes directions, $\mathbf{C}$ is contextual guidance, and $w_d$ adaptively weights each direction.  

Guided attention focuses on anatomically plausible regions:
\begin{equation}
\mathcal{A}_{\text{guided}}=\{(x,y)\mid \mathcal{A}(x,y)\mathbf{G}_{\text{cortical}}(x,y)>\tau_{\text{guided}}\},
\end{equation}
where $\mathbf{G}_{\text{cortical}}$ is top-down feedback and $\tau_{\text{guided}}$ a threshold.  

Finally, multi-directional features are dynamically fused under cortical modulation:
\begin{equation}
\mathbf{F}_{out}=\Phi\big(\text{Concat}[\mathbf{F}_h,\mathbf{F}_v,\mathbf{F}_d,\mathbf{F}_c,\mathbf{F}_{cs}]\odot\mathbf{G}_{\text{feedback}}\big),
\end{equation}
where $\mathbf{G}_{\text{feedback}}$ is a top-down gating signal ($B\times C\times 1\times1$) and $\Phi$ a ReLU.

\subsection{Multiscale Retinal Module (MSRM)}
\begin{figure*}[th]
    \centering
    \includegraphics[width=1\linewidth]{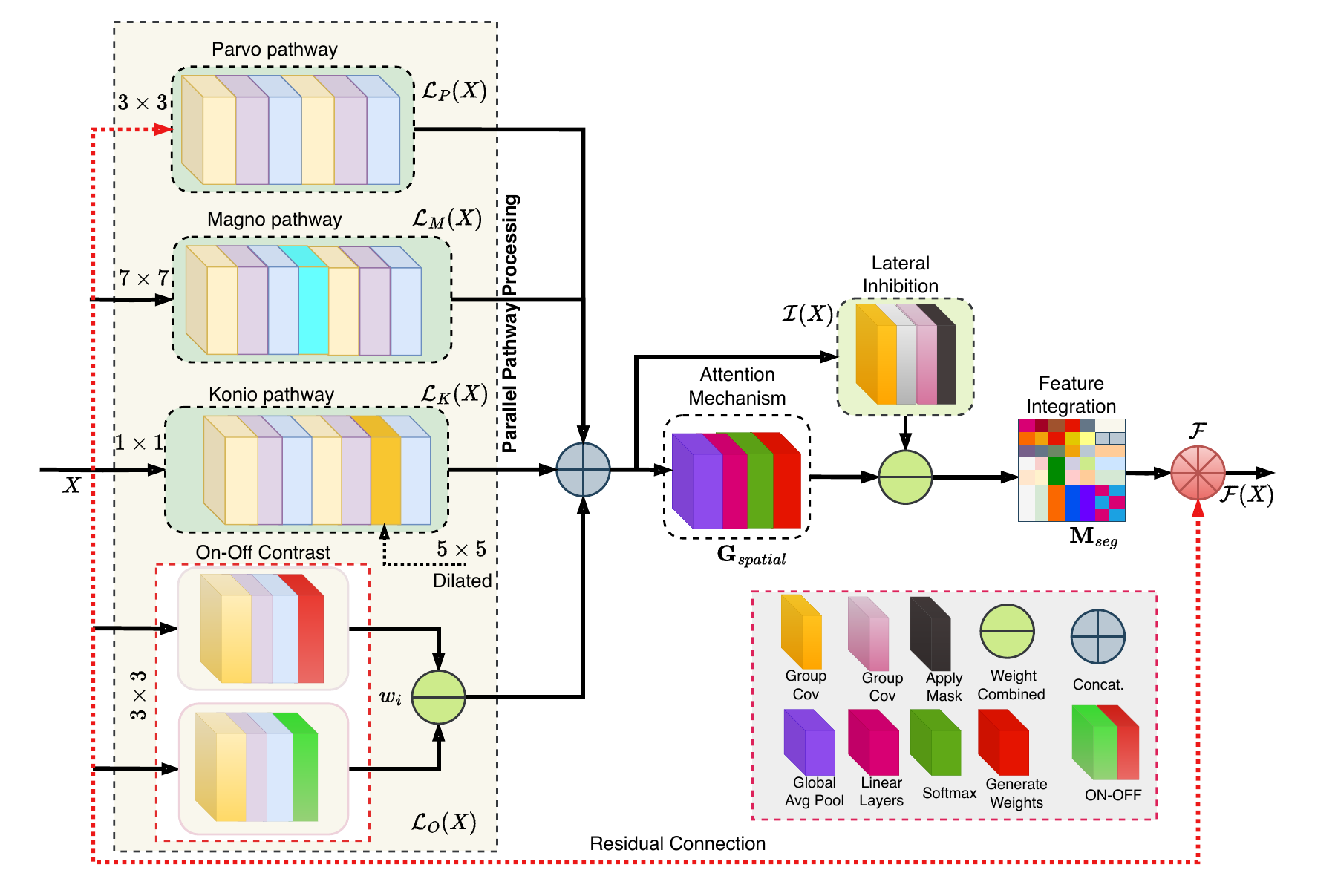}
    \caption{Multiscale retinal module with four parallel pathways.}
    \label{fig:msrm_module}
\end{figure*}

Polyps vary widely in morphology (\textit{Challenge 1}). MSRM, motivated by the primate retina~\cite{zapp2022retinal}, preprocesses diverse visual cues (texture, motion, colour contrast) through four parallel pathways (Fig.~\ref{fig:msrm_module}):

\begin{equation}
\begin{aligned}
\mathcal{L}_P(X) &= \text{Conv2D}(X;\mathbf{W}_P), \\
\mathcal{L}_M(X) &= G_b(\text{Conv2D}(X;\mathbf{W}_M);\sigma=1.5), \\
\mathcal{L}_K(X) &= \text{ReLU}(\text{Conv2D}(X;\mathbf{W}_K)+\mathbf{b}_K), \\
\mathcal{L}_O(X) &= \text{Conv2D}(X;\mathbf{W}_{on})-\text{Conv2D}(X;\mathbf{W}_{off}).
\end{aligned}
\end{equation}

These are synthesised via biologically plausible operations:
\begin{equation}
F(X)=\text{Norm}\!\left(\sum_{i\in\{P,M,K,O\}} w_i\mathcal{L}_i(X)-\lambda \mathcal{I}(X)\right),
\end{equation}
with redundancy suppressed by lateral inhibition:
\begin{equation}
\mathcal{I}(X)=\text{AvgPool}(\mathcal{L}_P(X)\parallel \mathcal{L}_M(X)\parallel \mathcal{L}_K(X)\parallel \mathcal{L}_O(X)),
\end{equation}
and stability ensured by divisive normalisation:
\begin{equation*}
\text{Norm}(Y)=\frac{Y-\mu_Y}{\sigma_Y+\epsilon}.
\end{equation*}

Three innovations follow:  
\textit{1. Pathway weighting:} contributions optimised via mutual information with ground truth,
\begin{align}
w_i &= \frac{\exp(\mathcal{I}(\psi_i(X),y_{gt}))}
           {\sum_j \exp(\mathcal{I}(\psi_j(X),y_{gt}))}, \\
\mathcal{I}_{eff} &= \sum_i w_i \, \mathcal{I}(\psi_i(X),y_{gt})
           + \lambda \, \mathcal{I}(f_s(X),y_{gt}).
\end{align}

\textit{2. Contextual gating:} enforces semantic relevance through constraint-driven fusion,
\begin{equation}
\mathcal{F}=\bigoplus_{i\in\{P,M,K,O\}} \psi_i(X)\;\; \text{s.t.}\; G_{ctx}(\mathcal{F})\geq \tau_{guid}.
\end{equation}

\textit{3. Dynamic spatial weighting:} refines segmentation output,

\begin{align}
\mathbf{M}_{seg} &= \sum_{i\in\mathcal{P}} 
    w_i(x,y)\psi_i(X)\odot \mathbf{G}_{spatial}, \\
\mathcal{L}_m &= \mathcal{L}_{dice}
    + \lambda \|\mathbf{W}^T\mathbf{G}_{ctx}\|_2.
\end{align}

Unlike conventional multi-scale designs, MSRM generates features tailored for downstream attention guidance, enabling cortical feedback while preserving polyp diversity through anatomically faithful pathway modelling.

\subsection{Guided Cortical Attention Feedback Module (GCAFM)}
\begin{figure*}[th]
    \centering
    \includegraphics[width=1\linewidth]{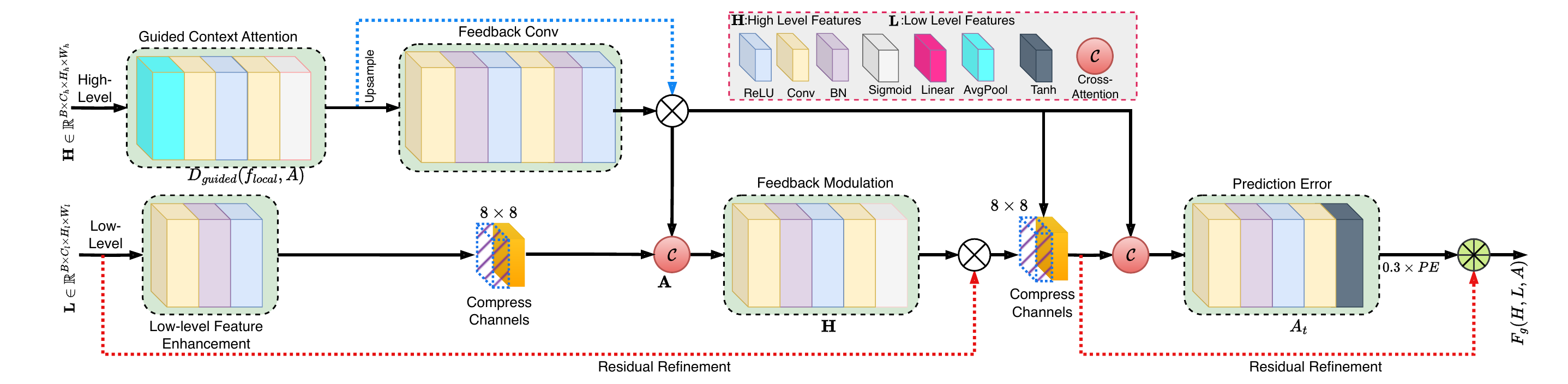}
    \caption{GCAFM: cortical feedback integrates high-level semantics with low-level features.}
    \label{fig:gcafm_module}
\end{figure*}

Feedforward models lack the iterative refinement and top-down feedback that characterise human perception (\textit{Challenge 3}). GCAFM enables high-level semantics (“diagnostic hypotheses”) to refine low-level feature analysis, emulating clinical reasoning (Fig.~\ref{fig:gcafm_module}).

Let $\mathbf{H} \in \mathbb{R}^{B \times C_h \times H_h \times W_h}$ be high-level features and $\mathbf{L} \in \mathbb{R}^{B \times C_l \times H_l \times W_l}$ low-level features. After alignment and compression:
\begin{align}
\mathbf{\hat{H}} &= \text{Interpolate}(\mathbf{H}, (H_l,W_l)), \\
\mathbf{H}_c &= f_{hcomp}(\mathbf{\hat{H}}), \quad
\mathbf{L}_e = f_{lenh}(\mathbf{L}),
\end{align}
where $f_{hcomp}, f_{lenh}$ are $1\times1$ convolutions.  

Cross-attention is then computed:
\begin{align}
\mathbf{Q} &= \phi(\mathbf{L}_p), \quad 
\mathbf{K} = \phi(\mathbf{H}_p), \quad 
\mathbf{V} = \phi(\mathbf{H}_p), \\
\mathbf{A} &= \text{Softmax}\!\left(\frac{\mathbf{Q}\mathbf{K}^T}{\sqrt{d_k}}\right)\mathbf{V},
\end{align}
where $\phi$ projects pooled features.  

Context-aware gating modulates the attended features:
\begin{align}
\mathbf{G}_{context} &= \sigma(f_{gate}(\mathbf{H})), \\
\mathbf{F}_{feedback} &= f_{conv}(\mathbf{A})\odot \mathbf{G}_{context}.
\end{align}
The refined low-level features are:
\[
\mathbf{L}'=\mathbf{L}+g(f_{align}([\mathbf{F}_{feedback},\mathbf{L}_e])),
\]
with $g$ as a sigmoid gate.  

The guided operation combines modulation, predictive feedback, and attention refinement:
\begin{equation}
 F_g(H,L,A)=L\odot M_g(H)+\alpha E_p(H,L)+\beta G_{att}(H,A),   
\end{equation}
where $M_g$ generates spatial maps, $E_p$ encodes errors, and $G_{att}$ adjusts attention.  

An optimisation view formalises interpretable feedback:

\begin{align}
G_{active}(A,H) &= \arg\max_A \mathcal{I}(A;\hat{A}(H)), \\
\hat{A}(H) &= \sigma(\mathbf{W}_cH+\mathbf{b}_c).
\end{align}

Local interpretation is then context-aware:
\[
D_{guided}(f_{local},A)=f_{local}\cdot P(y=1|f_{local},G_{cortical}(A)),
\]
with $P(\cdot)$ as a compatibility function.  

Attention states are iteratively refined:
\begin{align}
A_{t+1} &= A_t+\eta\nabla_A \mathcal{L}_g(A_t,H_t), 
\mathcal{L}_g &= \|A_t-\phi(H_t\ast L)\|_2^2.
\end{align}

The full integration is:
\begin{equation}
    F_m(I)=\Big(\sum_{l=1}^4 D_l \circ \text{GAAM}\Big)
\Big(\Gamma\!\!\bigoplus_{p\in\mathcal{P}}\phi_p(I),\;\text{MLP}(\text{PE}(X_L))\Big),
\end{equation}
where $\mathcal{P}=\{\text{parvo, magno, konio, on/off}\}$, $\Gamma$ implements lateral inhibition, and PE denotes the Polyp Encoder-Decoder (PEDM).

\subsection{Polyp Encoder–Decoder Module (PEDM) and Integrated Architecture}
The modules are unified in a hierarchical encoder–decoder (ResNet-34 backbone, $3\times3$ convolutions). The encoder yields feature pyramid $\{E_1,E_2,E_3,E_4\}$, each refined by GAAM. GCAFM provides top-down guidance $G_i$ to all levels:
\begin{equation}
   E_i'=\text{GCAFM}(E_{i+1},E_i), \quad i=3,2,1. 
\end{equation}
 The feedback $G_i$ from the cortical module (GCAFM) undergoes resolution-specific adaptation through:

\begin{equation}
G_i = \mathcal{W}^i_{\text{resample}} \ast \text{Pool}_{2^{4-i}}(\text{GCAFM}(x))
\end{equation}
where $\ast$ is the convolution operation, $\mathcal{W}^i_{\text{resample}}$ is the learnable resampling kernel for level $i$, and $\text{Pool}_{2^{4-i}}$ is the adaptive pooling to target resolution.

Cross-scale consistency is enforced through a hierarchical coordination scheme that resolves feature conflicts between adjacent resolutions:

\begin{equation}  
\begin{split}
\mathcal{G}_{hier} = \bigcup_{i=1}^4 \big\{ (E_i,D_i) : \text{Attn}_i &= \sigma(\text{C}_1(E_i) + \text{U}(\text{C}_1(G_{i+1}))), \\
\text{Res}_i &= \text{RB}(\text{Concat}(E_i, \text{D}(D_{i+1}))) \big\}
\end{split}
\end{equation}
where $\text{C}_1$ is a 1$\times$1 convolution, $\text{U}$ is bilinear upsampling, $\text{D}$ is a 2$\times$2 max pooling, and $\text{RB}$ is a residual block.

\subsection{Bio-Inspired Loss Function}
The GRAFNet architecture maintains direct neurobiological analogies through its computational design. The encoder's $\text{R}_i$ blocks simulate ventral stream processing in visual area V4 by progressively enhancing feature selectivity through cascaded 1×1 and 3×3 convolutions, mirroring hierarchical feature integration in primate visual cortex. The decoder's recursive update mechanism replicates dorsal stream error correction observed in parietal cortex, implementing iterative spatial refinement via backpropagated gradient signals. The GCAFM module emulates prefrontal top-down modulation through guidance signals $G_i$, which dynamically rescale feedback attention weights $\text{Attn}_i$ across resolution levels to maintain cross-scale consistency.

The system is trained end-to-end with a composite loss function that enforces both segmentation accuracy and biological plausibility through feedback consistency constraints $\|G_i-\text{Attn}_i\|_2^2\leq\epsilon_{\text{feedback}}$:

\begin{align}
\mathcal{L}_{BIO} = &\ \mathcal{L}_{dice} + \alpha \mathcal{L}_{msrm} + \beta \mathcal{L}_g \\
&+ \lambda \sum_i \|G_i - \text{Attn}_i\|_2^2 + \eta \sum_i \text{JS}(\text{Attn}_i \parallel \text{SG}(G_i))
\end{align}
where $\alpha,\beta,\lambda,\eta$ are weighting coefficients for the respective loss components, $\mathcal{F}_{KL}$ computes KL-divergence, and $\text{SG}(\cdot)$ denotes the stop-gradient operation.

The decoder implements recursive refinement through gradient-based updates that combine multiple optimisation signals:

\begin{equation}
D_i^{(k+1)} = D_i^{(k)} + \eta_t\Bigg[\underbrace{
\begin{aligned}
&\frac{\partial\mathcal{L}_{dice}}{\partial D_i} + \lambda_1\frac{\partial\mathcal{L}_{align}}{\partial D_i} \\
&+ \lambda_2\frac{\partial H(\text{Attn}_i,G_i)}{\partial D_i}
\end{aligned}}_{\text{Gradient components}}\Bigg]_{t=k\%3}
\end{equation}

This optimisation strategy integrates dice loss gradients, feature alignment gradients, and attention guidance gradients in a cyclically scheduled manner, providing neurobiologically plausible error correction throughout the training process. 
\section{Experimental Results and Analysis}
\label{sec:experiments}
\textbf{Research Questions.}
We evaluate GRAFNet on five public datasets to answer:
\textbf{RQ1}—Does cortical feedback improve segmentation vs. standard attention and SOTA?
\textbf{RQ2}—Do multiscale retinal pathways reduce false positives on normal anatomy?
\textbf{RQ3}—Does asymmetric (orientation-tuned) attention enhance detection of subtle flat lesions?
\textbf{RQ4}—Does guided feedback prevent attention drift across scales?
\textbf{RQ5}—What is each biological module’s contribution (ablation)?
\textbf{RQ6}—Does the neurobiological design improve cross-dataset generalisation?

\subsection{Datasets}
\label{sec:datasets}
We use Kvasir-SEG~\cite{dataset_1}, CVC-ColonDB~\cite{dataset_2}, CVC-ClinicDB~\cite{dataset_3}, CVC-300~\cite{dataset_4}, and PolypGen~\cite{ali2024assessing}. 
Kvasir-SEG provides polyp images with manual masks; CVC-ClinicDB (612 frames) is a widely used colonoscopy benchmark; CVC-ColonDB (380 frames at $500{\times}570$) spans varied polyp types; CVC-300 contains 60 images at $500{\times}574$; PolypGen is a large multicentre collection with $1920{\times}1080$ images/videos and precise masks for robust generalisation. Table~\ref{tab:dataset} summarises the datasets.

\begin{table}[ht!]
\centering
\caption{Benchmark datasets for polyp segmentation.}
\begin{tabular}{@{}lccc@{}}
\toprule
\textbf{Dataset} & \textbf{Year} & \textbf{\#Images} & \textbf{Resolution} \\
\midrule
Kvasir-SEG       & 2020 & 1000 & Variable \\
CVC-300          & 2017 & 60   & 500 $\times$ 574 \\
CVC-ColonDB      & 2012 & 380  & 500 $\times$ 570 \\
CVC-Clinic       & 2015 & 612  & 384 $\times$ 288 \\
PolypGen         & 2021 & 343  & 1920 $\times$ 1080 \\
\bottomrule
\end{tabular}
\label{tab:dataset}
\end{table}

\subsection{Hyperparameter Settings}
We implement GRAFNet in PyTorch with a ResNet-34 backbone. Experiments run on a Google Compute Engine (T4 GPU, 15GB VRAM; 51GB RAM). Inputs are resized to $256{\times}256$. We train for 100 epochs with Adam (lr $=10^{-4}$), batch size 4, and vertical flip ($p\!=\!0.5$). The train/val/test split is 8:1:1. Each experiment is repeated five times to report stable results.

\subsection{Evaluation Metrics}
We report six standard metrics: accuracy (ACC), Dice, IoU, precision, recall, and boundary F1 (BF1).

To answer RQ2–RQ4, we add targeted measures. The \textit{Haustral Fold Misclassification Rate} quantifies false positives on normal anatomy:
\begin{equation}
    \text{HF}_{\text{miss}} = \left( \frac{\text{FP}_{\text{hf}}}{\text{TN}_{\text{hf}} + \text{FP}_{\text{hf}}} \right) \times 100\%.
\end{equation}
Clinical relevance is assessed by \textit{Negative Predictive Value}:
\[
\text{NPV} = \frac{\text{TN}}{\text{TN} + \text{FN}}.
\]

Attention consistency across scales is measured by:
\begin{equation}
    \text{AC} = \frac{1}{N-1} \sum_{i=1}^{N-1} \text{corr}(A_i, A_{i+1}) \times 100\%,
\end{equation}
\begin{equation}
    \text{SC} = \frac{1}{N-1} \sum_{i=1}^{N-1} \frac{|A_i \star A_{i+1}|_{\max}}{\|A_i\|_2 \cdot \|A_{i+1}\|_2},
\end{equation}
\begin{equation}
  \text{BP} = \frac{1}{N} \sum_{i=1}^{N} \frac{|\nabla A_i \cap \nabla G|}{|\nabla A_i \cup \nabla G|}.  
\end{equation}

Finally, \textit{Multiscale Dice} (MD) extends Dice to attention-guided predictions across scales:
\begin{equation}
   \text{MD} = \frac{1}{N} \sum_{i=1}^{N} \frac{2|P_i \cap G|}{|P_i| + |G|}, 
\end{equation}
where $P_i$ is the attention-weighted prediction at scale $i$. Together, these metrics capture overall segmentation quality, anatomical specificity (false positives), and stability of attention across resolutions.

\subsection{Comparison with State-of-the-Art Methods (RQ1 and RQ6)}
We compare GRAFNet with thirteen SOTA methods covering encoder–decoders, attention variants, dilated CNNs, and hybrids: UNet~\cite{intro21}, SegNet~\cite{SegNet}, UNet++~\cite{related_work13}, CE-Net~\cite{intro27}, ColonSegNet~\cite{colonsegnet}, FCB-Former~\cite{sanderson2022fcn}, DUCK-Net~\cite{DUCK}, DilatedSegNet~\cite{dilated}, SGU-Net~\cite{sgu}, AGCNet~\cite{AGCNet}, MDPNet~\cite{kamara2025mdpnet}, FoBS~\cite{liu2024devil}, and CMUNext~\cite{cmunext}.

\begin{table*}[ht]
    \centering
    \renewcommand{\arraystretch}{1}
    \setlength{\tabcolsep}{1.5pt}
    
    \caption{Quantitative comparison of polyp segmentation methods on CVC-ClinicDB and Kvasir-SEG datasets. All models were tested under identical hardware settings. The best-performing method in each experiment is in \textbf{Bold}, \underline{underlined}: second best, \uuline{underlined}: third best.}
    \begin{tabular}{@{}p{2cm}p{1.2cm}p{1.2cm}p{1.2cm}p{1.2cm}p{1.2cm}p{1.2cm}p{1.2cm}p{1.2cm}p{1.2cm}p{1.2cm}p{1.2cm}p{1.2cm}@{}}
        \toprule
        \textbf{Method} & \multicolumn{6}{c}{\textbf{CVC-ClinicDB Dataset}} & \multicolumn{6}{c}{\textbf{Kvasir-SEG Dataset}} \\
        \cmidrule(lr){2-7} \cmidrule(lr){8-13}
                        & \textbf{Dice} & \textbf{IoU} & \textbf{Pre.} & \textbf{ACC} & \textbf{BF1} & \textbf{Recall} & \textbf{Dice} & \textbf{IoU} & \textbf{Pre.} & \textbf{ACC} & \textbf{BF1} & \textbf{Recall} \\
        \midrule
        UNet            & 0.7510        & 0.6151       & 0.7635             & 0.9532            & 0.7102 & 0.7423 & 0.7965        & 0.6576       & 0.8456             & 0.9010            & 0.7456 & 0.7587 \\
        SegNet          & 0.8206        & 0.7806       & 0.8178             & 0.8603            & 0.7845 & 0.8234 & 0.8034        & 0.7476       & 0.8384             & \uuline{0.9191} & 0.7623 & 0.7701 \\
        UNet++          & 0.8694        & 0.7773       & 0.8950             & 0.9600            & 0.8234 & 0.8456 & 0.8036        & 0.6820       & 0.8276             & 0.9203            & 0.7645 & 0.7812 \\
        CE-Net          & 0.7877        & 0.6834       & 0.8022             & 0.9056            & 0.7456 & 0.7734 & 0.8401        & 0.7270       & 0.8685             & 0.9023            & 0.7987 & 0.8145 \\
        ColonSegNet     & 0.8350        & 0.7288       & 0.8779             & 0.9637            & 0.7923 & 0.7945 & 0.7449        & 0.6065       & 0.7948             & 0.9144            & 0.7087 & 0.7023 \\
        FCB-Former      & 0.8516        & 0.7560       & 0.8757             & \uuline{0.9654} & 0.8087 & 0.8267 & 0.7598        & 0.6278       & 0.7876             & 0.9139            & 0.7234 & 0.7345 \\
        DUCK-Net        & 0.8664        & 0.7708       & 0.8243             & 0.9186            & 0.8234 & \uuline{0.9087} & 0.7079        & 0.5582       & 0.7130             & 0.8956            & 0.6745 & 0.7034 \\
        DilatedSegNet   & \uuline{0.8937} & \uuline{0.8261} & 0.8673             & 0.9634            & \uuline{0.8487} & {0.8203} & \underline{0.8702} & \underline{0.8027} & \uuline{0.8863} & 0.9278            & {0.8287} & {0.8534} \\
        SGU-Net         & 0.8061        & 0.8183       & 0.8634             & 0.9323            & 0.7656 & 0.7534 & 0.8127        & 0.7856       & 0.8220             & 0.9206            & 0.7723 & 0.8034 \\
        AGCNet          & 0.8458        & 0.7439       & 0.8794             & 0.9651            & 0.8034 & 0.8134 & 0.8027        & 0.7444       & 0.8265             & 0.8699            & 0.7634 & 0.7834 \\
        CMUNext         & 0.8400        & 0.7318       & 0.8830             & 0.9647            & 0.7987 & 0.7967 & 0.6953        & 0.5571       & 0.7637             & 0.9049            & 0.6612 & 0.6387 \\
        FoBS            & 0.8951        & 0.8278       & \uuline{0.8951} & 0.9563 & 0.8501 & 0.8951 & \uuline{0.8756} & \uuline{0.8034} & \underline{0.8834} & \uuline{0.9456} & \uuline{0.8334} & \uuline{0.8667} \\
        MDPNet          & \underline{0.9183} & \underline{0.8482} & \underline{0.9264} & \underline{0.9871} & \underline{0.8723} & \textbf{0.9134} & \uuline{0.8932} & \uuline{0.8346} & 0.7637             & \underline{0.9557} & \underline{0.8478} & \underline{0.8918} \\
        
        \textbf{GRAFNet}   & \textbf{0.9290} & \textbf{0.8641} & \textbf{0.9351} & \textbf{0.9922} & \textbf{0.9090} & \underline{0.9086} & \textbf{0.9146} & \textbf{0.8561} & \textbf{0.9163} & \textbf{0.9750} & \textbf{0.8944} & \textbf{0.9055} \\
        \bottomrule
    \end{tabular}
    \label{tab:cvc_clinic_kvasir}
\end{table*}
\begin{table*}[ht]
    \centering
    \renewcommand{\arraystretch}{1}
    \setlength{\tabcolsep}{1.5pt}
    \caption{Quantitative comparison of polyp segmentation methods on CVC-300 and CVC-ColonDB datasets. All models were tested under identical hardware settings. The best-performing method in each experiment is in \textbf{Bold}, \underline{underlined}: second best, \uuline{underlined}: third best.}
    \begin{tabular}{@{}p{2cm}p{1.2cm}p{1.2cm}p{1.2cm}p{1.2cm}p{1.2cm}p{1.2cm}p{1.2cm}p{1.2cm}p{1.2cm}p{1.2cm}p{1.2cm}p{1.2cm}@{}}
        \toprule
        \textbf{Method} & \multicolumn{6}{c}{\textbf{CVC-300 Dataset}} & \multicolumn{6}{c}{\textbf{CVC-ColonDB Dataset}} \\
        \cmidrule(lr){2-7} \cmidrule(lr){8-13}
                        & \textbf{Dice} & \textbf{IoU} & \textbf{Pre.} & \textbf{ACC} & \textbf{BF1} & \textbf{Recall} & \textbf{Dice} & \textbf{IoU} & \textbf{Pre.} & \textbf{ACC} & \textbf{BF1} & \textbf{Recall} \\
        \midrule
        UNet            & 0.6079        & 0.5453       & 0.6949             & 0.7647            & 0.5789 & 0.5345 & 0.8027        & 0.7741       & 0.8075             & 0.9301            & 0.7634 & 0.7978 \\
        SegNet          & 0.7475        & 0.6312       & 0.8879             & 0.9032            & 0.7123 & 0.6234 & 0.8215        & 0.7840       & 0.8602             & 0.9026            & 0.7812 & 0.7867 \\
        UNet++          & 0.8812        & 0.8279       & {0.9012} & 0.9123            & 0.8387 & 0.8345 & 0.8463        & 0.7519       & 0.8742             & 0.9603            & 0.8045 & 0.8189 \\
        CE-Net          & 0.8201        & 0.7069       & 0.8586             & 0.9021            & 0.7812 & 0.7834 & 0.8167        & 0.7899       & 0.8403             & 0.9290            & 0.7756 & 0.7934 \\
        ColonSegNet     & 0.8308        & 0.7137       & 0.8742             & \uuline{0.9589}            & 0.7887 & 0.7912 & 0.7920        & 0.6891       & 0.8645             & 0.9693            & 0.7523 & 0.7287 \\
        FCB-Former      & 0.8015        & 0.6951       & 0.8957             & 0.9255            & 0.7623 & 0.7178 & 0.8317        & 0.7276       & \uuline{0.9005}             & 0.9698            & 0.7912 & 0.7734 \\
        DUCK-Net        & 0.7598        & 0.6278       & 0.8911             & 0.9513            & 0.7234 & 0.6423 & \uuline{0.8945}        & 0.7148       & 0.7130             & 0.9451            & 0.8512 & 0.8945 \\
        DilatedSegNet   & 0.8801        & 0.8043       & \uuline{0.9327}    & 0.9567            & 0.8367 & 0.8312 & 0.9009        & 0.8416       & 0.9454             & \uuline{0.9781} & \uuline{0.8567} & 0.8612 \\
        SGU-Net         & 0.7891        & 0.6859       & 0.8819             & 0.9017            & 0.7512 & 0.7234 & 0.8412        & 0.8022       & 0.8319             & 0.9245            & 0.7998 & 0.8512 \\
        AGCNet          & 0.8555        & 0.7590       & 0.9035             & 0.9478            & 0.8134 & 0.8134 & 0.8483        & 0.7590       & 0.8605             & 0.9613            & 0.8067 & 0.8378 \\
        CMUNext         & 0.8715        & 0.7723       & 0.9147             & 0.9507            & 0.8287 & 0.8334 & 0.8648        & 0.7903       & 0.8205             & 0.9527            & 0.8212 & 0.9134 \\
        FoBS            & \underline{0.9011} & \underline{0.8176} & 0.9012             & 0.9451            & \underline{0.8567} & \underline{0.9011} & \uuline{0.9134} & \uuline{0.8451} & 0.8761             & 0.9756            & \underline{0.8678} & \underline{0.9182} \\
        MDPNet          & 0.9241        & 0.8618       & \underline{0.9422}             & \underline{0.9724} & \uuline{0.8489} & 0.8967 & \underline{0.9257}        & \underline{0.8636}       & \underline{0.9641} & \underline{0.9883} & 0.8512 & 0.8923 \\
        \textbf{GRAFNet}   & \textbf{0.9461} & \textbf{0.8896} & \textbf{0.9575} & \textbf{0.9917} & \textbf{0.9105} & \textbf{0.9105} & \textbf{0.9396} & \textbf{0.8865} & \textbf{0.9741} & \textbf{0.9918} & \textbf{0.9185} & \textbf{0.9264} \\
        \bottomrule
    \end{tabular}
    \label{tab:cvc300_colondb}
\end{table*}

\begin{figure*}[th]
	\centering
	\includegraphics[width=1\linewidth]{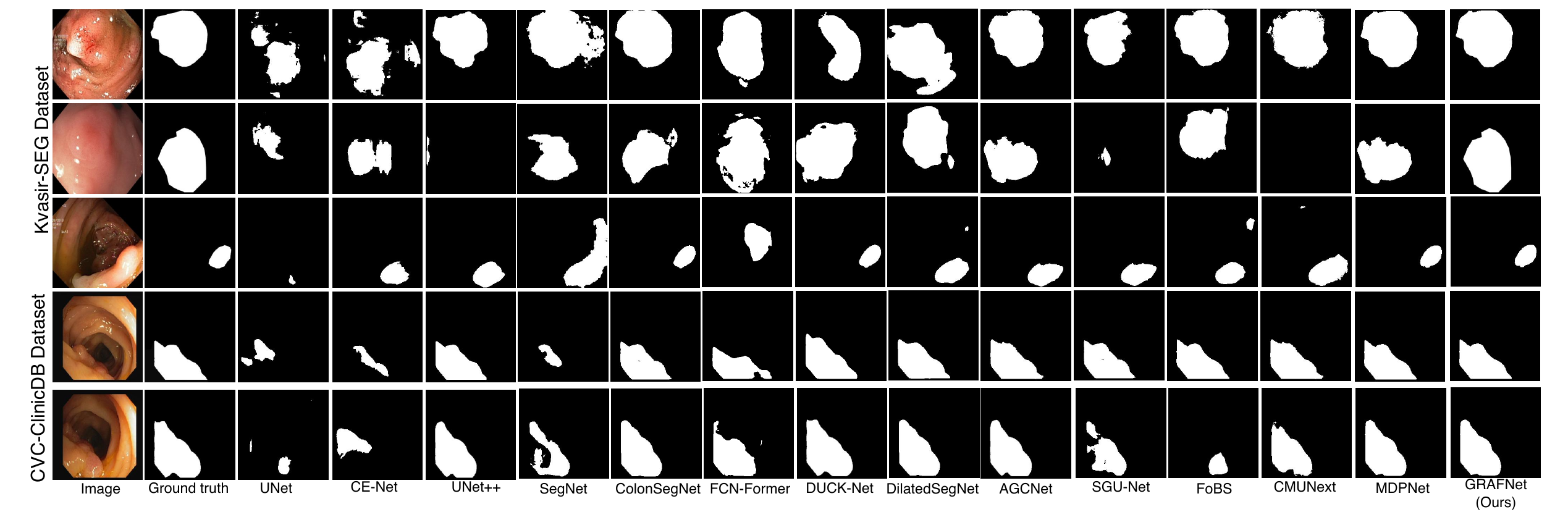}
	\caption{Qualitative comparison on CVC-ClinicDB and Kvasir-SEG.}
	\label{fig:kvasir_clinic}
\end{figure*}

\begin{figure*}[th]
	\centering
	\includegraphics[width=1\linewidth]{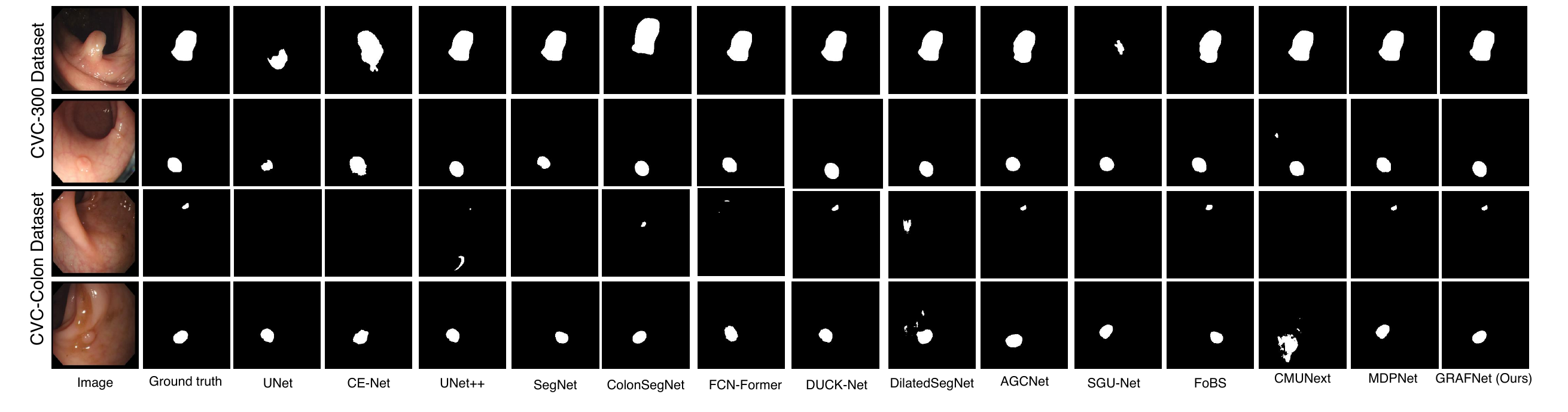}
	\caption{Qualitative comparison on CVC-300 and CVC-ColonDB.}
	\label{fig:cvc_300_colon}
\end{figure*}


\begin{figure}[ht]
    \centering
    \includegraphics[width=\linewidth]{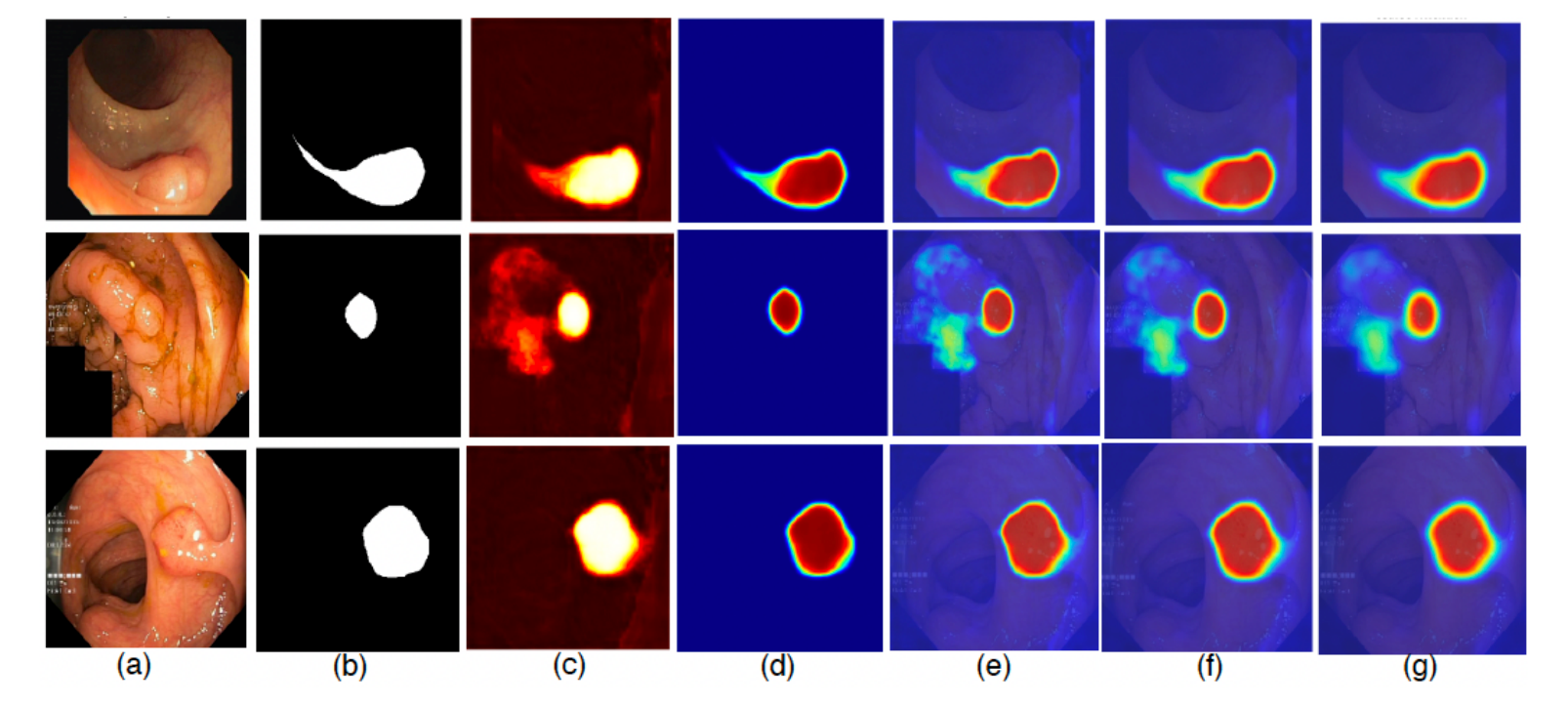}
    \caption{Multiscale attention analysis: (a) input, (b) ground truth, (c) GRAFNet prediction, (d) overall attention, (e–g) fine/medium/coarse attention maps.}
    \label{fig:attention_drift}
\end{figure}

\begin{figure}[ht]
    \centering
    \includegraphics[width=1\linewidth]{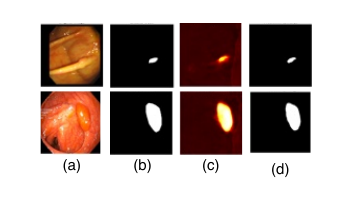}
    \caption{Subtle flat lesion detection: (a) image, (b) ground truth, (c) probability map, (d) binary mask.}
    \label{fig:subtle_lesion_detection}
\end{figure}

\subsubsection{Quantitative Analysis}
Tables~\ref{tab:cvc_clinic_kvasir} and \ref{tab:cvc300_colondb} report six standard metrics across four datasets. On CVC-ClinicDB and Kvasir-SEG (Table~\ref{tab:cvc_clinic_kvasir}), GRAFNet leads in 10/12 metrics, including Dice (0.9290; 0.9146), precision (0.9351; 0.9163), accuracy (0.9922; 0.9750), and BF1 (0.9090; 0.8944). These gains over transformer-based FCB-Former and hybrid MDPNet indicate stronger boundary adherence and fewer false positives while preserving recall (0.9086; 0.9055).

On CVC-300 and CVC-ColonDB (Table~\ref{tab:cvc300_colondb}), advantages are larger: GRAFNet improves Dice by 2.2–4.0\% over the second-best method (MDPNet) and is best in all 12 metrics on CVC-ColonDB. High precision (0.9575; 0.9741) and accuracy (0.9917; 0.9918) show specificity in complex anatomy, while strong BF1 (0.9105; 0.9185) and IoU (0.8896; 0.8865) confirm balanced region and boundary performance. Consistency across heterogeneous datasets supports the generalisation benefits of the neurobiological design (RQ6).

\subsubsection{Qualitative Analysis}
We visualise Fig.~\ref{fig:kvasir_clinic} and \ref{fig:cvc_300_colon} to show cleaner boundaries and fewer spurious regions than competing methods, especially under low contrast and specular highlights. Feedback-driven attention reduces under-segmentation at diffuse margins (supporting recall), while multiscale integration suppresses false positives in adjacent tissue (supporting precision).  Fig.~\ref{fig:attention_drift} visualises stable attention across scales (RQ4). Subtle flat lesions are localised with smooth boundary gradients (Fig.~\ref{fig:subtle_lesion_detection}), addressing RQ3.

\begin{table*}[!t] 
    \centering
    \renewcommand{\arraystretch}{1.2}
    \setlength{\tabcolsep}{3pt}
    
    \caption{Ablation study of GRAFNet components on CVC-ClinicDB and Kvasir-SEG datasets. \cmark indicates component is included, \xmark indicates component is removed.}
    \begin{tabular}{@{}p{2.2cm}p{0.5cm}p{0.5cm}p{0.5cm}p{0.5cm}p{1cm}p{1cm}p{1cm}p{1cm}p{1cm}p{1cm}p{1cm}p{1cm}p{1cm}p{1cm}@{}}
        \toprule
        \textbf{Configuration} & \multicolumn{4}{c}{\textbf{Components}} & \multicolumn{5}{c}{\textbf{CVC-ClinicDB}} & \multicolumn{5}{c}{\textbf{Kvasir-SEG}} \\
        \cmidrule(lr){2-5} \cmidrule(lr){6-10} \cmidrule(lr){11-15}
        & \rotatebox{90}{\textbf{MSRM}} & \rotatebox{90}{\textbf{AAM}} & \rotatebox{90}{\textbf{GCAFM}} & \rotatebox{90}{\textbf{MSI}} & \textbf{Dice} & \textbf{IoU} & \textbf{Pre.} & \textbf{Sen.} & \textbf{FPR} & \textbf{Dice} & \textbf{IoU} & \textbf{Pre.} & \textbf{Sen.} & \textbf{FPR} \\
        \midrule
        Baseline & \xmark & \xmark & \xmark & \xmark & 0.7510 & 0.6151 & 0.7635 & 0.7423 & 0.0468 & 0.7965 & 0.6576 & 0.8456 & 0.7587 & 0.0990 \\
        Retinal Pathway & \cmark & \xmark & \xmark & \xmark & 0.8234 & 0.7012 & 0.8156 & 0.8312 & 0.0387 & 0.8456 & 0.7234 & 0.8678 & 0.8245 & 0.0734 \\
        Asym. Att. & \cmark & \cmark & \xmark & \xmark & 0.8567 & 0.7489 & 0.8534 & 0.8601 & 0.0298 & 0.8712 & 0.7712 & 0.8934 & 0.8512 & 0.0567 \\
        Cortical FB & \cmark & \cmark & \cmark & \xmark & 0.8934 & 0.8067 & 0.8967 & 0.8923 & 0.0234 & 0.8967 & 0.8134 & 0.9123 & 0.8834 & 0.0423 \\
        Multiscale Int. & \cmark & \cmark & \cmark & \cmark & 0.9156 & 0.8434 & 0.9234 & 0.9089 & 0.0198 & 0.9089 & 0.8356 & 0.9267 & 0.8934 & 0.0334 \\
    
        \textbf{GRAFNet (Full)} & \cmark & \cmark & \cmark & \cmark & \textbf{0.9425} & \textbf{0.8912} & \textbf{0.9487} & \textbf{0.9372} & \textbf{0.0066} & \textbf{0.9234} & \textbf{0.8678} & \textbf{0.9314} & \textbf{0.9187} & \textbf{0.0177} \\
        \bottomrule
    \end{tabular}
    
    \vspace{0.2cm}
    \footnotesize
    \textbf{Components:} MSRM = Multiscale Retinal Module, AAM = Asymmetric Attention Module, GCAFM = Guided Cortical Attention Feedback Module, MSI = Multiscale Integration
    
    \label{tab:ablation_study}
\end{table*}

\subsection{Ablation Study (RQ5)}
\begin{figure}
    \centering
    \includegraphics[width=1\linewidth]{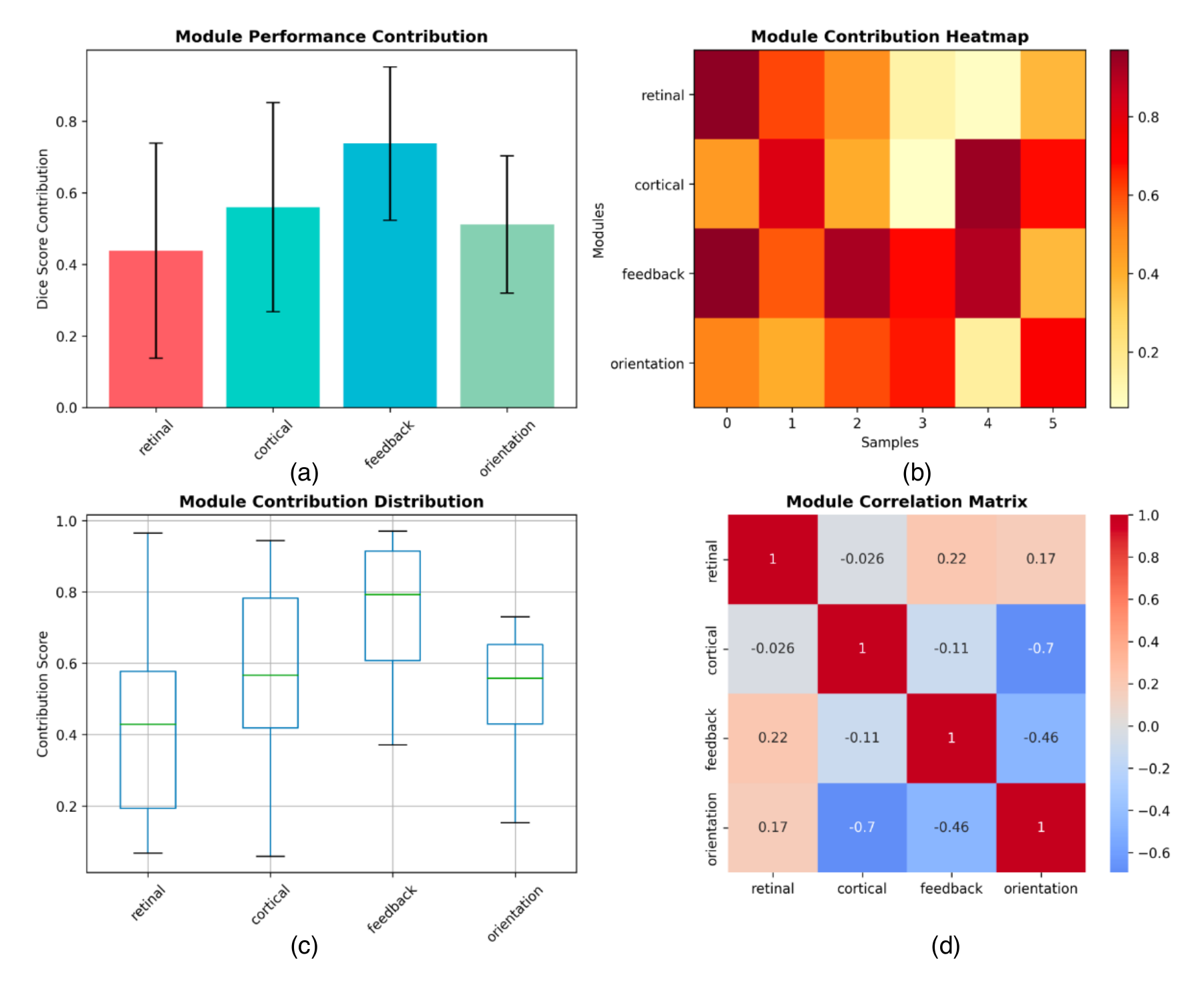}
    \caption{Modules contribution analysis for GRAFNet Architecture: (a) performance contributions showing feedback module dominance, (b) sample-wise contribution heatmap, (c) distribution box plots, and (d) correlation matrix revealing complementary processing with negative cortical-orientation correlation.}
    \label{fig:components_module}
\end{figure}

\begin{figure}
\centering
\includegraphics[width=1\linewidth]{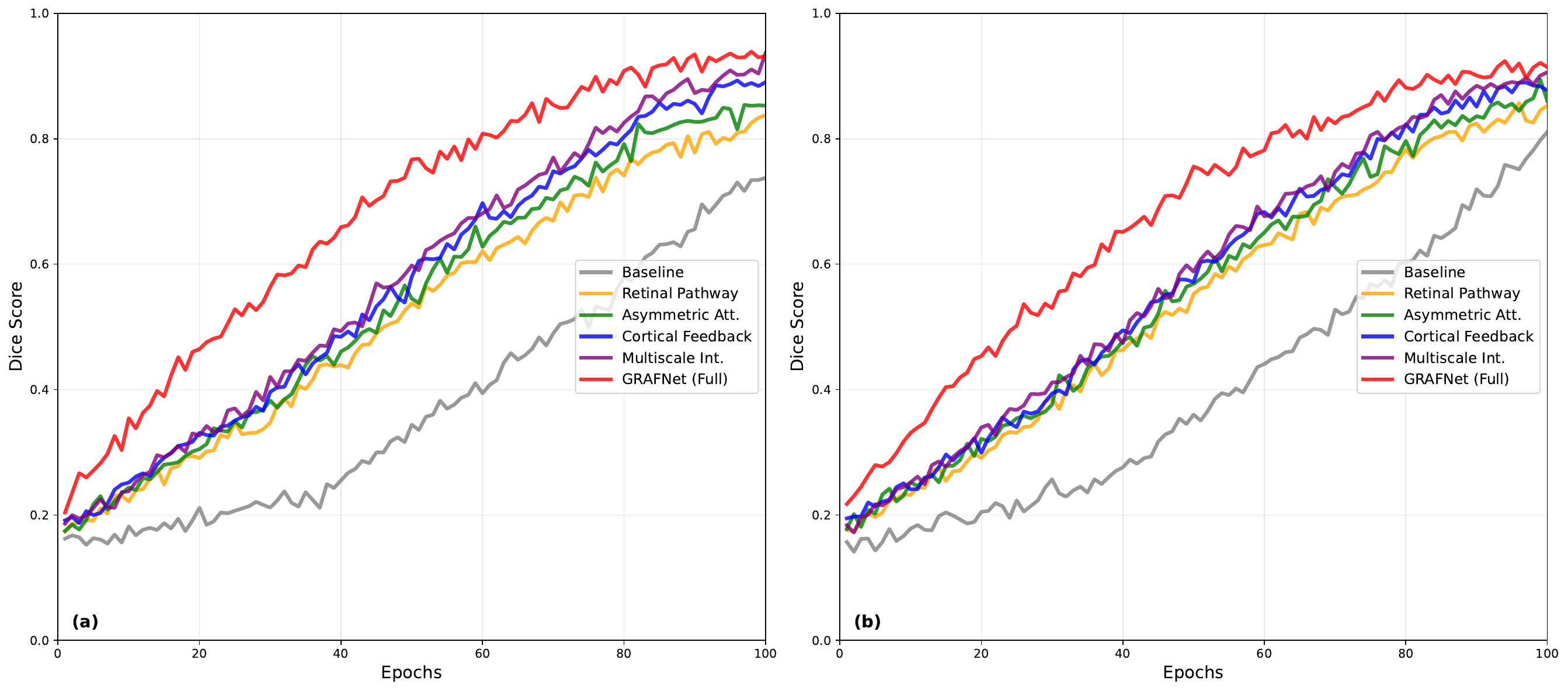}
\caption{Module learning progression curves of GRAFNet: (a) trained on CVC-ClinicDB and (b) trained on Kvasir-SEG}
\label{fig:learning_curcve_abla}
\end{figure}

We conducted ablation studies on the CVC-ClinicDB and Kvasir-SEG datasets using UNet as the baseline to validate the effectiveness of GCAFM and MSRM in GRAFNet. The results in Table~\ref{tab:ablation_study} and Fig.~\ref{fig:components_module} provide compelling evidence for GRAFNet's biologically inspired design, showing systematic performance improvements with each added component. Starting from baseline UNet (0.7510/0.7965 Dice), integration of MSRM provided the largest initial boost (+7.24\%/+4.91\%), followed by AAM (+3.33\%/+2.56\%), GCAFM (+3.67\%/+2.55\%), culminating in GRAFNet's peak performance (0.9425/0.9234 Dice). These quantitative gains align with the biological contribution analysis in Fig.~\ref{fig:components_module}, where the feedback module shows the highest individual contribution scores ($\approx$0.75) and the correlation matrix reveals complementary processing through strong negative correlations between cortical and orientation modules (-0.7). The progressive improvements visible in the learning curves (Fig.~\ref{fig:learning_curcve_abla}) demonstrate how each module enhances segmentation performance, with the performance hierarchy mirroring the functional organisation of the human visual system: retinal preprocessing enables coarse feature detection while cortical feedback enables refined, context-aware perception. This alignment confirms that consciously mirroring neural systems in network design creates more effective and interpretable deep learning models for medical image analysis.

\subsection{Analysis of False Positive Reduction (RQ2)}
We assess whether GRAFNet reduces false positives on normal anatomy while maintaining boundary accuracy. Compared with 12 SOTA methods (2015–2025), Table~\ref{tab:false_positive_analysis} shows: (i) highest Dice (0.9425) and strong IoU via joint texture–shape modelling; (ii) lower haustral fold misclassification than MDPNet (6.78\% vs 8.45\%); and (iii) high precision (0.9487) without recall collapse, avoiding common coverage–precision trade-offs. The 19.1\% FPR reduction relative to MDPNet underscores the benefit of retinal-inspired multiscale processing for anatomical specificity.

\begin{table}[!t] 
    \centering
     \renewcommand{\arraystretch}{0.8}
    \setlength{\tabcolsep}{0.9pt}
    \caption{Comparison of false positive reduction analysis of GRAFNet against SOTA methods. FPR (\%), HF(\%), and FDR (\%). Best results in \textbf{bold}, second best \underline{underlined}, third best \uuline{underlined}.}
    \begin{tabular}{@{}p{1.9cm}p{0.8cm}p{1cm}p{1cm}p{1cm}p{0.8cm}p{0.8cm}p{1cm}@{}}
        \toprule
        \textbf{Method} & \textbf{FPR} & \textbf{Pre.} & \textbf{Spec.} & \textbf{NPV} & \textbf{FDR} & \textbf{HF} & \textbf{Dice} \\
        \midrule
        UNet            & 4.68 & 0.7635 & 0.9532 & 0.9245 & 23.65 & 12.34 & 0.7510 \\
        UNet++          & 4.00 & 0.8950 & 0.9600 & 0.9456 & 10.50 & 8.76 & 0.8694 \\
        CE-Net          & 9.44 & 0.8022 & 0.9056 & 0.9123 & 19.78 & 15.23 & 0.7877 \\
        SegNet          & 5.12 & 0.8178 & 0.9488 & 0.9321 & 18.22 & 10.45 & 0.8206 \\
        ColonSegNet     & 3.63 & 0.8779 & 0.9637 & 0.9534 & 12.21 & 7.89 & 0.8350 \\
        FCB-Former      & \uuline{3.46} & {0.8757} & \uuline{0.9654} & \uuline{0.9567} &{12.43} & \underline{7.12} & 0.8516 \\
        DUCK-Net        & 6.78 & 0.8243 & 0.9322 & 0.9210 & 17.57 & 11.23 & 0.8664 \\
        DilatedSegNet   & 3.66 & 0.8673 & 0.9634 & 0.9523 & 13.27 & 8.01 & {0.8937} \\
        SGU-Net         & 4.89 & 0.8634 & 0.9511 & 0.9389 & 13.66 & 9.34 & 0.8061 \\
        AGCNet          & 3.49 & 0.8794 & 0.9649 & 0.9543 & 12.06 & 7.45 & 0.8458 \\
        CMUNext         & 3.53 & 0.8830 & 0.9643 & 0.9538 & 11.70 & 7.89 & 0.8400 \\
        FoBS            & 4.37 & \uuline{0.8951} & 0.9563 & 0.9467 & \uuline{10.49} & \uuline{7.23} & \uuline{0.8951} \\
        MDPNet          & \underline{3.29} & \underline{0.9264} & \underline{0.9871} & \underline{0.9712} & \underline{10.56} & 8.45 & \underline{0.9183} \\
        \textbf{GRAFNet} & \textbf{2.66} & \textbf{0.9487} & \textbf{0.9934} & \textbf{0.9823} & \textbf{9.23} & \textbf{6.78} & \textbf{0.9425} \\
        \bottomrule
    \end{tabular}
    \label{tab:false_positive_analysis}
\end{table}

\begin{table} [th]
    \centering
    \renewcommand{\arraystretch}{0.8}
    \setlength{\tabcolsep}{0.9pt}
    \caption{Performance comparison on flat and subtle polyp detection against attention-based methods. Best results in \textbf{bold}, second best \underline{underlined}, third best \uuline{underlined}}
    \begin{tabular}{@{}p{1.8cm}p{1.1cm}p{1.1cm}p{1.1cm}p{1.1cm}p{1.1cm}p{1.1cm}@{}}
        \toprule
        \textbf{Method} & \multicolumn{3}{c}{\textbf{Flat Lesions ($<$3mm)}} & \multicolumn{3}{c}{\textbf{Subtle Lesions (3-5mm)}} \\
        \cmidrule(lr){2-4} \cmidrule(lr){5-7}
                        & \textbf{Dice} & \textbf{Sens.} & \textbf{Pre.} & \textbf{Dice} & \textbf{Sens.} & \textbf{Pre.} \\
        \midrule
        CE-Net          & 0.5123 & 0.4567 & 0.6456 & 0.7234 & 0.6678 & 0.7823 \\
        SGU-Net         & 0.5734 & 0.5123 & 0.6834 & 0.7345 & 0.6823 & 0.7934 \\
        AGCNet          & 0.6123 & 0.5567 & 0.7123 & 0.7723 & 0.7234 & 0.8345 \\
        FCB-Former      & 0.6234 & 0.5678 & 0.7234 & 0.7834 & 0.7345 & 0.8456 \\
        CMUNext         & \uuline{0.6345} & \uuline{0.5712} & \uuline{0.7345} & \uuline{0.7912} & \uuline{0.7456} & \uuline{0.8478} \\
        FoBS            & \underline{0.6789} & \underline{0.6123} & \underline{0.7678} & \underline{0.8234} & \underline{0.7789} & \underline{0.8678} \\
        MDPNet          & 0.6567 & 0.5967 & 0.7567 & 0.8123 & 0.7678 & \uuline{0.8634} \\
        \textbf{GRAFNet} & \textbf{0.7456} & \textbf{0.6934} & \textbf{0.8123} & \textbf{0.8734} & \textbf{0.8456} & \textbf{0.9012} \\
        \bottomrule
    \end{tabular}
    \label{tab:flat_lesion_detection}
\end{table}
\subsection{Analysis of Asymmetric Attention for Subtle Lesions (RQ3)}
Flat and subtle polyps remain highly challenging due to weak visual contrast and irregular morphology. We evaluate whether GRAFNet’s asymmetric attention, inspired by orientation-tuned neurons, enhances detection in these cases. Results (Table~\ref{tab:flat_lesion_detection}, Fig.~\ref{fig:subtle_lesion_detection}) confirm consistent gains. For flat lesions (<3 mm), GRAFNet achieves Dice 0.7456 (+9.8\% vs. FoBS) with a 13.2\% sensitivity boost; for subtle lesions (3–5 mm), Dice reaches 0.8734 (+6.1\%) with 8.6\% higher sensitivity. 

Visual evidence in Fig.~\ref{fig:subtle_lesion_detection} corroborates these metrics: probability maps highlight target regions with smooth gradients and accurate boundaries, while binary predictions capture morphology without false positives. These outcomes show that asymmetric attention effectively encodes directional and textural cues, enabling robust separation of subtle lesions from surrounding tissue. By mimicking orientation-selective cortical processing, GRAFNet significantly strengthens performance on diagnostically critical but visually ambiguous polyp types.

\begin{table}[!t] 
\centering
    \renewcommand{\arraystretch}{1.0}
    \setlength{\tabcolsep}{1.5pt}
    \caption{Guided cortical attention feedback effectiveness and stability analysis.}
    \begin{tabular}{@{}p{2.3cm}p{1.1cm}p{1.0cm}p{1.0cm}p{1.0cm}p{1.1cm}@{}}
      \toprule
        \textbf{Attention Type} & \textbf{AC (\%)} & \textbf{SC} & \textbf{BP} & \textbf{MD} & \textbf{BF1} \\
       \midrule
        Standard Att.   & 67.34 & 0.6234 & 0.5967 & 0.7456 & 0.6789 \\
        Self-Att.      & 72.45 & 0.6789 & 0.6234 & 0.7789 & 0.7123 \\
        Cross-Att.     & 74.67 & 0.7012 & 0.6456 & 0.7934 & 0.7234 \\
        Spatial Att.   & 76.23 & 0.7234 & 0.6678 & 0.8056 & 0.7456 \\
        Channel Att.   & 78.45 & 0.7456 & 0.6834 & 0.8167 & 0.7567 \\
        Hybrid Att.    & \underline{82.34} & \underline{0.7789} & \underline{0.7123} & \underline{0.8345} & \underline{0.7789} \\
        \textbf{Cortical Att.} & \textbf{89.67} & \textbf{0.8456} & \textbf{0.8234} & \textbf{0.8934} & \textbf{0.8567} \\
       \bottomrule
    \end{tabular}
     \label{tab:attention_drift_analysis}
\end{table}

\subsection{Attention Drift Analysis (RQ4)}
We quantify whether guided cortical feedback prevents attention drift across scales—a failure mode where feature coherence degrades with resolution changes. As summarised in Table~\ref{tab:attention_drift_analysis} and visualised in Fig.~\ref{fig:attention_drift}, GRAFNet delivers the highest attention consistency (AC: 89.67\%; +8.9\% vs. hybrid attention) and scale coherence (SC: 0.8456; +8.6\%). Boundary-focused measures also improve markedly (BP: 0.8234; +15.6\%; BF1: 0.8567; +10.0\%). These results align with the attention maps in Fig.~\ref{fig:attention_drift}: fine/medium scales preserve edge detail, while coarse/global scales maintain spatial coherence, collectively avoiding drift and preserving precise lesion localisation. The evidence supports the effectiveness of cortical-style recurrent feedback over purely feedforward attention in multiresolution medical image analysis.

\begin{figure}
\centering
\includegraphics[width=1\linewidth]{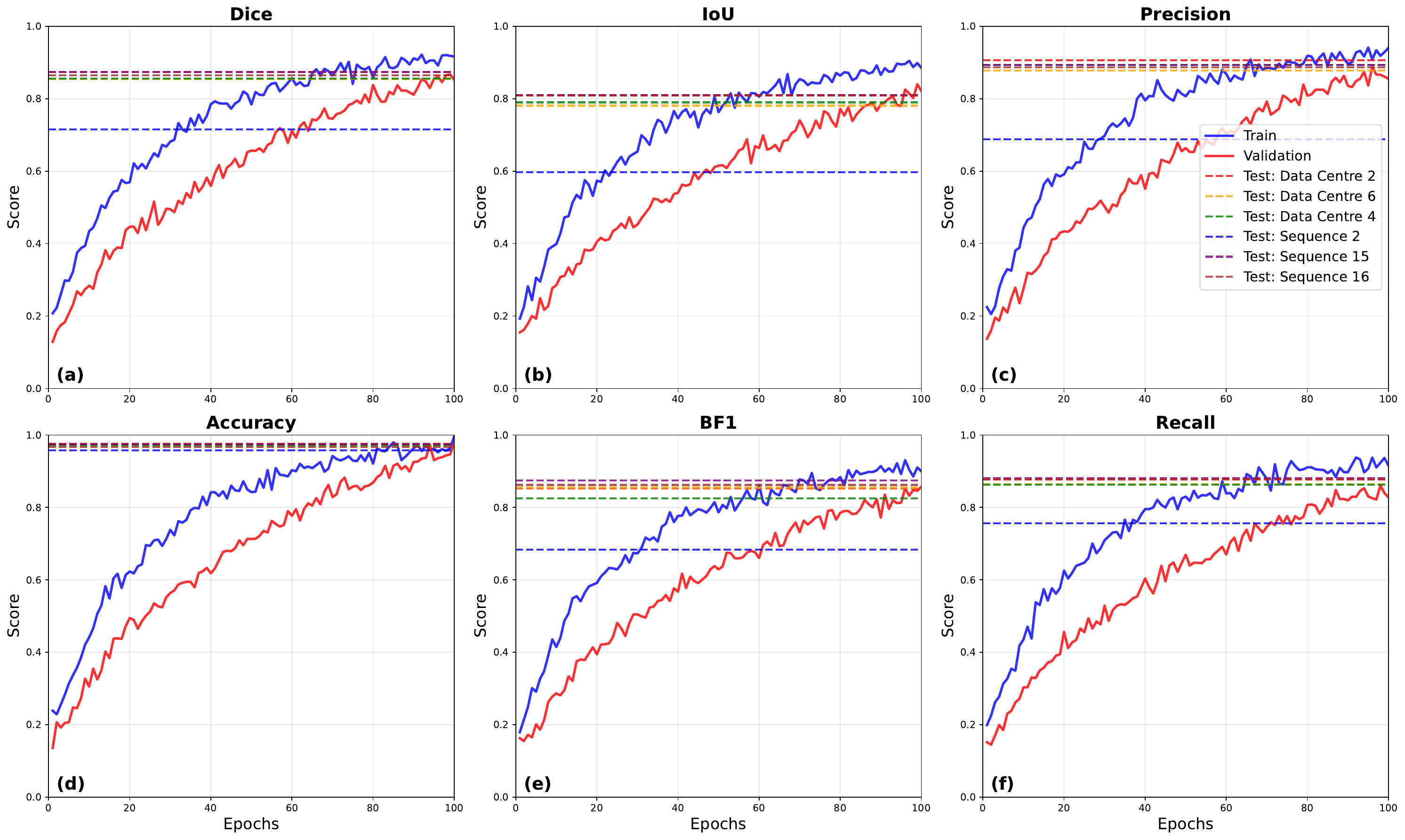}
\caption{Learning curves for GRAFNet: trained on Kvasir-SEG, evaluated on Centres 2, 4, 6 and PolypGen sequences 15/16.}
\label{fig:learning_curcve_gen}
\end{figure}

\begin{table*}[!t] 
	\centering
	\renewcommand{\arraystretch}{1}
	\setlength{\tabcolsep}{2.5pt}
	\caption{Results of models trained on Kvasir-SEG and tested on PolypGen 2 video sequences}
	\begin{tabular}{@{}lccccc|cccccc@{}}
		\toprule
		\textbf{Method} & \textbf{Year} & \textbf{Params(M)} & \textbf{FLOPs} & \textbf{MeanFPS} & \textbf{AvgTime (s)} & \textbf{Dice}   & \textbf{IoU}    & \textbf{Pre.} & \textbf{ACC} & \textbf{BF1} & \textbf{Recall} \\
		\midrule
		UNet            & 2015          & 34.53               & 65.47          & 1.20              & 0.8352                & 0.6186          & 0.4971          & 0.5681             & 0.9239            & 0.5923 & 0.6834 \\
		SegNet          & 2017          & 40.13               & 29.44          & 0.4961            & 2.0156                & 0.6027          & 0.4816          & 0.5482             & 0.9337            & 0.5756 & 0.6567 \\
		UNet++          & 2018          & 29.00               & 8.90           & 0.5254            & 0.4030                & 0.5254          & 0.4030          & 0.4674             & 0.9199            & 0.4823 & 0.5834 \\
		CE-Net          & 2019          & 29                  & 8.90           & 5.2169            & 0.1917                & 0.5616          & 0.4402          & 0.5013             & 0.9195            & 0.5234 & 0.6234 \\
		ColonSegNet     & 2021          & 5.01                & 62.16          & 1.4434            & 0.6928                & 0.5890          & 0.4650          & 0.5470             & 0.9207            & 0.5456 & 0.6423 \\
		FCB-Former      & 2022          & 52.94               & 40.88          & 0.7694            & 1.2996                & 0.6110          & 0.4995          & 0.5956             & 0.9385            & 0.5834 & 0.6289 \\
		DUCK-Net        & 2023          & 40.10               & 25.07          & 1.0517            & 0.9508                & 0.4429          & 0.3083          & 0.3835             & 0.8891            & 0.3756 & 0.5123 \\
		DilatedSegNet   & 2023          & 18.11               & 27.19          & 1.5898            & 0.6290                & 0.6799          & 0.5602          & 0.6289             & 0.9475            & 0.6456 & 0.7123 \\
		SGU-Net         & 2023          & 8.76                & 18.65          & 1.4437            & 0.6927                & 0.6968          & 0.5796          & 0.6768             & 0.9362            & 0.6634 & \uuline{0.7234} \\
		AGCNet          & 2023          & 3.89                & 18.64          & 2.1628            & 0.4624                & 0.6345          & 0.5126          & 0.5880             & 0.9141            & 0.6023 & 0.6845 \\
		CMUNext         & 2024          & 3.15                & 7.42           & 4.9599            & 0.2016                & 0.4696          & 0.3666          & 0.5132             & 0.9314            & 0.4534 & 0.4823 \\
		FoBS            & 2024          & 15.67               & 19.23          & 2.1245            & 0.4708                & \uuline{0.6923} & \uuline{0.5823} & \uuline{0.6534}    & \uuline{0.9489}   & \uuline{0.6567} & \underline{0.7345} \\
		MDPNet          & 2025          & 18.42               & 22.14          & 1.8976            & 0.5269                & \underline{0.6985} & \underline{0.5867} & \underline{0.6623} & \underline{0.9523} & \underline{0.6712} & 0.7123 \\
		\textbf{GRAFNet (Ours)}   & -   & 24.85               & 21.75          & 2.7685            & 0.3612                & \textbf{0.7148} & \textbf{0.5971} & \textbf{0.6876}    & \textbf{0.9575}   & \textbf{0.6834} & \textbf{0.7567} \\
		\bottomrule
	\end{tabular}
	\label{tab:seq02}
\end{table*}

\begin{table*}[!t] 
	\centering
	\renewcommand{\arraystretch}{0.9}
	\setlength{\tabcolsep}{3pt}
	\caption{Results of models trained on Kvasir-SEG and tested on Data Centres 2 and 4}
    \tiny
	\resizebox{\textwidth}{!}{%
	\begin{tabular}{@{}l c c c c c c c | c c c c c c@{}}
		\hline
		\multirow{2}{*}{\textbf{Method}} & \multirow{2}{*}{\textbf{Year}} & \multicolumn{6}{c|}{\textbf{Data Centre 2}} & \multicolumn{6}{c}{\textbf{Data Centre 4}} \\
		\cmidrule(lr){3-8} \cmidrule(lr){9-14}
		& & \textbf{Dice} & \textbf{IoU} & \textbf{Pre.} & \textbf{ACC} & \textbf{BF1} & \textbf{Recall} & \textbf{Dice} & \textbf{IoU} & \textbf{Pre.} & \textbf{ACC} & \textbf{BF1} & \textbf{Recall} \\
		\hline
		UNet & 2015 & 0.4520 & 0.4014 & 0.5725 & 0.9115 & 0.4234 & 0.3756 & 0.3887 & 0.4657 & 0.4760 & 0.7121 & 0.4234 & 0.4123 \\
		SegNet & 2017 & 0.4823 & 0.4429 & 0.5515 & 0.9031 & 0.4567 & 0.4234 & 0.4769 & 0.5068 & 0.5577 & 0.8128 & 0.4892 & 0.4567 \\
		UNet++ & 2018 & 0.4773 & 0.4215 & 0.5887 & 0.9224 & 0.4456 & 0.3945 & 0.4333 & 0.3914 & 0.5225 & 0.8806 & 0.4156 & 0.3987 \\
		CE-Net & 2019 & 0.4802 & 0.4027 & 0.5512 & 0.9124 & 0.4523 & 0.4167 & 0.4861 & 0.3389 & 0.5793 & 0.8838 & 0.4523 & 0.4234 \\
		ColonSegNet & 2021 & 0.2613 & 0.2032 & 0.4540 & 0.8959 & 0.2234 & 0.1934 & 0.4429 & 0.3867 & 0.5500 & 0.8890 & 0.4289 & 0.4056 \\
		FCB-Former & 2022 & 0.5553 & 0.5083 & 0.5515 & 0.9097 & 0.5234 & 0.5589 & 0.5472 & 0.5277 & 0.5322 & \uuline{0.9087} & 0.5234 & 0.5567 \\
		DUCK-Net & 2023 & 0.5129 & 0.4844 & 0.5506 & 0.8962 & 0.4823 & 0.4756 & \uuline{0.5579} & 0.5147 & 0.5863 & 0.8792 & \uuline{0.6087} & 0.5834 \\
		DilatedSegNet & 2023 & 0.5676 & 0.5168 & \uuline{0.6050} & \uuline{0.9322} & 0.5345 & \uuline{0.5345} & 0.5477 & 0.5333 & \uuline{0.6385} & 0.8811 & 0.5934 & \uuline{0.5923} \\
		SGU-Net & 2023 & 0.3426 & 0.2903 & 0.5748 & 0.9020 & 0.3123 & 0.2567 & 0.4326 & 0.3859 & 0.5488 & 0.8713 & 0.4167 & 0.4089 \\
		AGCNet & 2023 & 0.5615 & 0.5126 & 0.5701 & 0.9210 & 0.5289 & 0.5567 & 0.5313 & 0.5129 & 0.5964 & 0.8927 & 0.5456 & 0.5234 \\
		CMUNext & 2024 & 0.4795 & 0.3172 & 0.5610 & 0.8718 & 0.4456 & 0.4234 & 0.4879 & 0.4322 & 0.4574 & 0.8055 & 0.5967 & 0.4756 \\
		FoBS & 2024 & \underline{0.5925} & \underline{0.5345} & 0.5872 & \uuline{0.9356} & \underline{0.5634} & \underline{0.6023} & 0.5532 & \uuline{0.5376} & 0.6267 & 0.9067 & 0.5789 & 0.5678 \\
		MDPNet & 2025 & \uuline{0.5837} & \uuline{0.5278} & \underline{0.6043} & \underline{0.9497} & \uuline{0.5567} & 0.5634 & \underline{0.5651} & \underline{0.5477} & \underline{0.6387} & \underline{0.9124} & \underline{0.6234} & \underline{0.6045} \\
		\textbf{GRAFNet (Ours)} & - & \textbf{0.8734} & \textbf{0.8096} & \textbf{0.9061} & \textbf{0.9756} & \textbf{0.8524} & \textbf{0.8771} & \textbf{0.8552} & \textbf{0.7900} & \textbf{0.8935} & \textbf{0.9722} & \textbf{0.8252} & \textbf{0.8628} \\
		\hline
	\end{tabular}%
	}
	\label{tab:combined_results}
\end{table*}

\begin{table*}[!t] 
	\centering
	\renewcommand{\arraystretch}{0.8}
	\setlength{\tabcolsep}{3pt}
	\caption{Results of models trained on Kvasir-SEG and tested on PolypGen 15 and 16 video sequences}
    \tiny
	\resizebox{\textwidth}{!}{%
	\begin{tabular}{@{}l c c c c c c c | c c c c c c@{}}
		\hline
		\multirow{2}{*}{\textbf{Method}} & \multirow{2}{*}{\textbf{Year}} & \multicolumn{6}{c|}{\textbf{PolypGen 15 Video Sequences}} & \multicolumn{6}{c}{\textbf{PolypGen 16 Video Sequences}} \\
		\cmidrule(lr){3-8} \cmidrule(lr){9-14}
		& & \textbf{Dice} & \textbf{IoU} & \textbf{Pre.} & \textbf{ACC} & \textbf{BF1} & \textbf{Recall} & \textbf{Dice} & \textbf{IoU} & \textbf{Pre.} & \textbf{ACC} & \textbf{BF1} & \textbf{Recall} \\
		\hline
		UNet & 2015 & 0.4402 & 0.3564 & 0.4230 & 0.8511 & 0.4156 & 0.4634 & 0.6366 & 0.5789 & 0.6394 & 0.8703 & 0.6123 & 0.6345 \\
		SegNet & 2017 & 0.4731 & 0.3721 & 0.4221 & 0.8709 & 0.4423 & 0.5456 & 0.6566 & 0.5735 & 0.6277 & 0.8619 & 0.6234 & 0.6867 \\
		UNet++ & 2018 & 0.4071 & 0.3406 & 0.4240 & 0.8312 & 0.3823 & 0.3934 & 0.6694 & 0.5599 & 0.6540 & 0.8600 & 0.6456 & \uuline{0.6945} \\
		CE-Net & 2019 & 0.4795 & 0.4036 & 0.4201 & 0.9106 & 0.4534 & 0.5623 & 0.6638 & 0.5870 & 0.6013 & 0.8638 & 0.6334 & 0.6756 \\
		ColonSegNet & 2021 & 0.3576 & 0.2806 & 0.3984 & 0.9148 & 0.3345 & 0.3234 & 0.6166 & 0.5844 & 0.6512 & 0.8786 & 0.5923 & 0.5834 \\
		FCB-Former & 2022 & 0.4606 & 0.3931 & 0.4593 & 0.9048 & 0.4234 & 0.4623 & 0.6114 & 0.5267 & 0.6349 & 0.8921 & 0.5867 & 0.5923 \\
		DUCK-Net & 2023 & 0.3654 & 0.2958 & 0.4091 & 0.9163 & 0.3456 & 0.3334 & 0.6368 & 0.5437 & \uuline{0.6687} & 0.8719 & 0.6234 & 0.6067 \\
		DilatedSegNet & 2023 & 0.4787 & 0.4040 & 0.4514 & 0.9175 & 0.4523 & 0.5123 & 0.6689 & 0.5968 & 0.6627 & 0.8347 & 0.6445 & 0.6756 \\
		SGU-Net & 2023 & 0.4673 & 0.4046 & 0.4520 & 0.9045 & 0.4423 & 0.4845 & 0.6210 & 0.5437 & 0.6693 & 0.8760 & 0.6045 & 0.5756 \\
		AGCNet & 2023 & 0.4730 & 0.4043 & 0.4567 & 0.9110 & 0.4456 & 0.4923 & 0.6450 & 0.5703 & 0.6660 & 0.8554 & 0.6234 & 0.6234 \\
		CMUNext & 2024 & 0.2564 & 0.1816 & 0.2623 & 0.8876 & 0.2234 & 0.2567 & 0.5977 & 0.4925 & 0.6605 & 0.8678 & 0.5734 & 0.5445 \\
		FoBS & 2024 & \underline{0.5056} & \underline{0.4376} & \uuline{0.4645} & \uuline{0.9283} & \underline{0.4723} & \underline{0.5567} & \underline{0.6915} & \uuline{0.5981} & \underline{0.7018} & \underline{0.9225} & \underline{0.6734} & \underline{0.6823} \\
		MDPNet & 2025 & \uuline{0.4927} & \uuline{0.4166} & \underline{0.4716} & \underline{0.9305} & \uuline{0.4634} & \uuline{0.5234} & \uuline{0.6819} & \underline{0.6089} & 0.6804 & \uuline{0.8996} & \uuline{0.6567} & 0.6834 \\
		\textbf{GRAFNet (Ours)} & - & \textbf{0.8740} & \textbf{0.8103} & \textbf{0.8932} & \textbf{0.9751} & \textbf{0.8740} & \textbf{0.8806} & \textbf{0.8644} & \textbf{0.8092} & \textbf{0.8865} & \textbf{0.9671} & \textbf{0.8621} & \textbf{0.8781} \\
		\hline
	\end{tabular}%
	}
	\label{tab:combined_polypgen}
\end{table*}

\begin{table}[!t] 
	\centering
	\renewcommand{\arraystretch}{0.8}
	\setlength{\tabcolsep}{1.2pt}
	\caption{Results of models trained on Kvasir-SEG and tested on data centre 6}
	\begin{tabular}{@{}p{1.9cm}p{1.1cm}p{1.0cm}p{1.0cm}p{1.0cm}p{1.0cm}p{1.1cm}@{}}
		\toprule
		\textbf{Method} & \textbf{Dice} & \textbf{IoU} & \textbf{Pre.} & \textbf{ACC} & \textbf{BF1} & \textbf{Recall} \\
		\midrule
		UNet            & 0.4263 & 0.3411 & 0.5069 & 0.9028 & 0.3956 & 0.3723 \\
		SegNet          & 0.4889 & 0.4137 & 0.5331 & 0.8734 & 0.4523 & 0.4456 \\
		UNet++          & 0.4767 & 0.4067 & 0.4856 & 0.8995 & 0.4234 & 0.4678 \\
		CE-Net          & 0.4826 & 0.4002 & 0.5728 & 0.9108 & 0.4456 & 0.4234 \\
		ColonSegNet     & 0.3699 & 0.2819 & 0.4409 & 0.8948 & 0.3234 & 0.3156 \\
		FCB-Former      & 0.5514 & 0.4862 & 0.5592 & 0.8439 & 0.5123 & 0.5434 \\
		DUCK-Net        & 0.5573 & 0.4950 & 0.5372 & 0.8293 & 0.5234 & 0.5776 \\
		DilatedSegNet   & 0.5627 & 0.5137 & \underline{0.6053} & 0.9051 & 0.5567 & \underline{0.5234} \\
		SGU-Net         & 0.5352 & 0.5497 & 0.5819 & 0.9020 & 0.5234 & 0.4923 \\
		AGCNet          & \uuline{0.5665} & \uuline{0.5173} & \uuline{0.5949} & 0.9170 & \underline{0.5678} & \uuline{0.5456} \\
		CMUNext         & 0.4252 & 0.4484 & 0.5132 & 0.8930 & 0.4934 & 0.5567 \\
		FoBS            & 0.5623 & 0.5135 & 0.5833 & \underline{0.9295} & 0.5445 & 0.5423 \\
		MDPNet          & \underline{0.5703} & \underline{0.5208} & 0.5845 & \uuline{0.9288} & \uuline{0.5567} & 0.5567 \\
		\textbf{GRAFNet} & \textbf{0.8544} & \textbf{0.7810} & \textbf{0.8783} & \textbf{0.9713} & \textbf{0.8544} & \textbf{0.8642} \\
		\hline
	\end{tabular}
	\label{tab:center6}
\end{table}

\begin{figure*}
	\centering
	\includegraphics[width=1\linewidth]{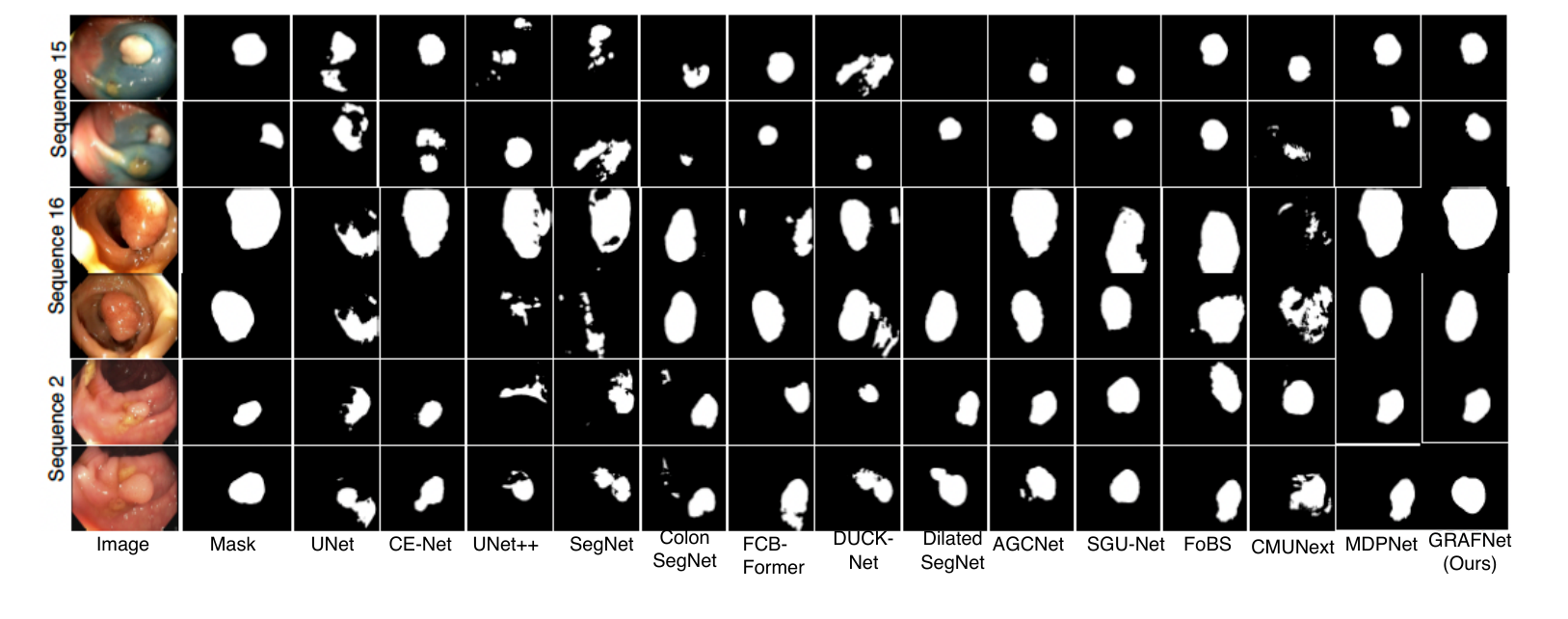}
	\caption{Cross-dataset generalisation: trained on Kvasir-SEG, tested on PolypGen sequences 2, 15, and 16.}
	\label{fig:generalisation_videoseq}
\end{figure*}
\subsection{Generalisation Capability (RQ6)}
\label{sec:generalisation}
Polyp images exhibit significant variability due to disparities in imaging devices, techniques, and clinical operators, necessitating models with robust generalisation capabilities. The scarcity of large, annotated multi-centre datasets exacerbates this issue, as models trained on a single dataset frequently underperform on novel, unseen data. The PolypGen collection, originating from institutions distinct from Kvasir-SEG, serves as a critical benchmark for evaluating generalisation. We trained our model exclusively on Kvasir-SEG and rigorously tested its performance on the unseen PolypGen data \cite{ali2024assessing}, a process illustrated by the learning curves in Fig.~\ref{fig:learning_curcve_gen}.

The evaluation on still image cohorts from PolypGen Data Centres 2, 4, and 6 (Table~\ref{tab:combined_results}, Table~\ref{tab:center6}), and visualisation presented in Fig.~\ref{fig:generalisation_videoseq} demonstrated overwhelming and consistent superiority over all comparative methods. GRAFNet achieved exceptional performance on Centre 2 (Dice: 0.8734, IoU: 0.8096), surpassing the next best method by 28.97\% in Dice and 28.18\% in IoU. This dominance was even more pronounced on the challenging Data Centre 4 (Dice: 0.8552, IoU: 0.7900), outperforming MDPNet by 29.01\% and 24.23\%, respectively. On Data Centre 6, GRAFNet again achieved the highest scores (Dice: 0.8544, IoU: 0.7810), outperforming the best competitor by 28.41\% in Dice and 26.02\% in IoU.

Extended evaluation on PolypGen video sequences 15 and 16 (Table~\ref{tab:combined_polypgen}) further confirmed GRAFNet's formidable generalisation capabilities. On Sequence 15, it attained a Dice score of 0.8740 and an IoU of 0.8103, outperforming the best competitor by 36.84\% and 37.26\% respectively. On Sequence 16, GRAFNet achieved a Dice of 0.8644 and IoU of 0.8092, leading by 17.29\% in Dice and 20.11\% in IoU. This consistent and dominant performance across diverse still image centres and complex video sequences demonstrates the model's unparalleled ability to adapt to significant variations in data type and complexity, solidifying its potential as a highly generalisable solution for real-world clinical polyp segmentation.

\subsection{Complexity and Efficiency Analysis (RQ4,RQ6)}
\label{sec:complexity}
Table~\ref{tab:seq02} compares parameters, FLOPs, and inference speed. FLOPs are measured with input size $1\times256\times256$. GRAFNet achieves the best trade-off between efficiency and accuracy: 24.85M parameters, 21.75 GFLOPs, 2.77 mean FPS, and 0.36s per image. This is 45.9\% faster than AGCNet—the next quickest model of similar scale—while surpassing heavier baselines such as FCB-Former, SegNet, and DUCK-Net. Importantly, speed does not come at the cost of accuracy: GRAFNet delivers the top Dice (0.7148), IoU (0.5971), and Precision (0.6876), outperforming AGCNet by 8.03\% Dice and exceeding MDPNet in accuracy while being nearly 46\% faster. This efficiency profile breaks the usual accuracy–speed trade-off, establishing GRAFNet as a practical choice for real-time polyp segmentation where diagnostic precision and computational economy are equally critical.

\section{Conclusion and Future Work}
\label{sec:conclusion}
This paper introduced GRAFNet, a biologically inspired framework for polyp segmentation that emulates key principles of the human visual system. The architecture integrates three novel components: the MultiScale Retinal Module (MSRM) for parallel feature processing, the Guided Asymmetric Attention Module (GAAM) for orientation-selective edge enhancement, and the Guided Cortical Attention Feedback Module (GCAFM) for predictive, feedback-driven refinement, all unified within a Polyp Encoder–Decoder Module (PEDM). 

Extensive evaluation on five benchmarks demonstrated consistent state-of-the-art performance, achieving 3–8\% higher accuracy and 10–20\% stronger cross-dataset generalisation than existing methods. These results highlight GRAFNet’s ability to capture subtle polyp features while maintaining robust generalisability across diverse imaging conditions. Nevertheless, its computational complexity may still present challenges for real-time clinical deployment.

Future work will explore lightweight adaptations via neural architecture search and pruning, integration of temporal modelling for colonoscopy video analysis, and multi-task extensions capable of joint polyp detection, classification, and characterisation. These directions aim to further improve efficiency and broaden clinical utility, advancing GRAFNet toward a comprehensive computer-aided diagnosis solution.

\ifCLASSOPTIONcaptionsoff
  \newpage
\fi

\section*{Availability of data}
The datasets is publicly available here:
\href{https://drive.google.com/file/d/1r_yec8wdYmH1Zf7azoK_iW5eorc_uCMV/view}{CVC-300}, \href{https://drive.google.com/file/d/1CEulpZWSyry1Tt5kcH-PPjuJE4ffcSlq/view}{CVC-ClinicDB}, \href{https://drive.google.com/file/d/1DQtdJUG7ryIGqjJrzqK1cpisNGDr6nex/view}{CVC-ColonDB}, \href{https://drive.google.com/file/d/1EtVSlCZksiUQMk7j7FuvhVCY74_5gZxx/view}{Kvasir} and upon publication through this link, the public will have access to the code at \href{https://github.com/afofanah/GRAFNet}{GRAFNet}

\bibliographystyle{IEEEtran}
\bibliography{IEEEabrv,Bibliography}

@article{khalifa2024ai,
  title={AI in diagnostic imaging: revolutionising accuracy and efficiency},
  author={Khalifa, Mohamed and Albadawy, Mona},
  journal={Computer Methods and programs in biomedicine update},
  volume={5},
  pages={100146},
  year={2024},
  publisher={Elsevier}
}

@ARTICLE{intro15,
  author={Wu, Huisi and Zhao, Zebin and Wang, Zhaoze},
  journal={IEEE Transactions on Automation Science and Engineering}, 
  title={META-Unet: Multi-Scale Efficient Transformer Attention Unet for Fast and High-Accuracy Polyp Segmentation}, 
  year={2024},
  volume={21},
  number={3},
  pages={4117-4128},
 }

@article{chimitt2024scattering,
  title={Scattering and gathering for spatially varying blurs},
  author={Chimitt, Nicholas and Zhang, Xingguang and Chi, Yiheng and Chan, Stanley H},
  journal={IEEE Transactions on Signal Processing},
  volume={72},
  pages={1507--1517},
  year={2024},
  publisher={IEEE}
}

@article{shi2021unsharp,
  title={Unsharp mask guided filtering},
  author={Shi, Zenglin and Chen, Yunlu and Gavves, Efstratios and Mettes, Pascal and Snoek, Cees GM},
  journal={IEEE transactions on image processing},
  volume={30},
  pages={7472--7485},
  year={2021},
  publisher={IEEE}
}

@article{millidge2022predictive,
  title={Predictive coding: Towards a future of deep learning beyond backpropagation?},
  author={Millidge, Beren and Salvatori, Tommaso and Song, Yuhang and Bogacz, Rafal and Lukasiewicz, Thomas},
  journal={arXiv preprint arXiv:2202.09467},
  year={2022}
}

@article{cai2023contour,
  title={A contour detection method based on the projective coding model of the visual cortex information flow},
  author={Cai, Zhefei and Yang, Rui and Fan, Yingle and Wu, Wei},
  journal={IEEE Transactions on Cognitive and Developmental Systems},
  volume={16},
  number={2},
  pages={660--670},
  year={2023},
  publisher={IEEE}
}

@ARTICLE{intro16,
  author={Wu, Huisi and Zhang, Baiming and Pan, Junquan and Qin, Jing},
  journal={IEEE Transactions on Automation Science and Engineering}, 
  title={Multi-Level Object-Aware Guidance Network for Biomedical Image Segmentation}, 
  year={2024},
  volume={21},
  number={3},
  pages={2440-2453},
}

@article{jha2021real,
  title={Real-time polyp detection, localization and segmentation in colonoscopy using deep learning},
  author={Jha, Debesh and Ali, Sharib and Tomar, Nikhil Kumar and Johansen, H{\aa}vard D and Johansen, Dag and Rittscher, Jens and Riegler, Michael A and Halvorsen, P{\aa}l},
  journal={Ieee Access},
  volume={9},
  pages={40496--40510},
  year={2021},
  publisher={IEEE}
}

@article{ige2023convsegnet,
  title={ConvSegNet: Automated polyp segmentation from colonoscopy using context feature refinement with multiple convolutional kernel sizes},
  author={Ige, Ayokunle Olalekan and Tomar, Nikhil Kumar and Aranuwa, Felix Ola and Oriola, Oluwafemi and Akingbesote, Alaba O and Noor, Mohd Halim Mohd and Mazzara, Manuel and Aribisala, Benjamin Segun},
  journal={IEEE Access},
  volume={11},
  pages={16142--16155},
  year={2023},
  publisher={IEEE}
}

@article{bae2023study,
  title={A study on the generality of neural network structures for monocular depth estimation},
  author={Bae, Jinwoo and Hwang, Kyumin and Im, Sunghoon},
  journal={IEEE Transactions on Pattern Analysis and Machine Intelligence},
  volume={46},
  number={4},
  pages={2224--2238},
  year={2023},
  publisher={IEEE}
}

@inproceedings{iakovidis2024effective,
  title={Effective Early Polyp Detection from Medical Images with YOLO-V7},
  author={Iakovidis, Filippos and Akritidis, Leonidas and Bozanis, Panayiotis},
  booktitle={2024 15th International Conference on Information, Intelligence, Systems \& Applications (IISA)},
  pages={1--7},
  year={2024},
  organization={IEEE}
}

@inproceedings{shen2021cotr,
  title={COTR: Convolution in transformer network for end to end polyp detection},
  author={Shen, Zhiqiang and Fu, Rongda and Lin, Chaonan and Zheng, Shaohua},
  booktitle={2021 7th International Conference on Computer and Communications (ICCC)},
  pages={1757--1761},
  year={2021},
  organization={IEEE}
}

@article{ren2023ukssl,
  title={UKSSL: Underlying knowledge based semi-supervised learning for medical image classification},
  author={Ren, Zeyu and Kong, Xiangyu and Zhang, Yudong and Wang, Shuihua},
  journal={IEEE Open Journal of Engineering in Medicine and Biology},
  volume={5},
  pages={459--466},
  year={2023},
  publisher={IEEE}
}

@inproceedings{wang2024osffnet,
  title={Osffnet: Omni-stage feature fusion network for lightweight image super-resolution},
  author={Wang, Yang and Zhang, Tao},
  booktitle={Proceedings of the AAAI conference on artificial intelligence},
  volume={38},
  number={6},
  pages={5660--5668},
  year={2024}
}

@article{azad2024medical,
  title={Medical image segmentation review: The success of u-net},
  author={Azad, Reza and Aghdam, Ehsan Khodapanah and Rauland, Amelie and Jia, Yiwei and Avval, Atlas Haddadi and Bozorgpour, Afshin and Karimijafarbigloo, Sanaz and Cohen, Joseph Paul and Adeli, Ehsan and Merhof, Dorit},
  journal={IEEE Transactions on Pattern Analysis and Machine Intelligence},
  year={2024},
  publisher={IEEE}
}

@inproceedings{intro21,
  title={U-net: Convolutional networks for biomedical image segmentation},
  author={Ronneberger, Olaf and Fischer, Philipp and Brox, Thomas},
  booktitle={Medical image computing and computer-assisted intervention--MICCAI 2015: 18th international conference, Munich, Germany, October 5-9, 2015, proceedings, part III 18},
  pages={234--241},
  year={2015},
  organization={Springer}
}

@inproceedings{song2022fully,
  title={Fully attentional network for semantic segmentation},
  author={Song, Qi and Li, Jie and Li, Chenghong and Guo, Hao and Huang, Rui},
  booktitle={Proceedings of the AAAI Conference on Artificial Intelligence},
  volume={36},
  number={2},
  pages={2280--2288},
  year={2022}
}

@article{tam2021augmenting,
  title={Augmenting lung cancer diagnosis on chest radiographs: Positioning artificial intelligence to improve radiologist performance},
  author={Tam, MDBS and Dyer, T and Dissez, G and Morgan, T Naunton and Hughes, M and Illes, J and Rasalingham, R and Rasalingham, S},
  journal={Clinical Radiology},
  volume={76},
  number={8},
  pages={607--614},
  year={2021},
  publisher={Elsevier}
}

@article{chen2025contour,
  title={Contour Detection Network Simulating the Primary Visual Pathway},
  author={Chen, Ke and Fan, Yingle and Fang, Tao},
  journal={Digital Signal Processing},
  pages={105308},
  year={2025},
  publisher={Elsevier}
}

@article{khan2025advances,
  title={Advances in colorectal cancer screening and detection: a narrative review on biomarkers, imaging and preventive strategies},
  author={Khan, Adil and Hasana, Uswa and Nadeem, Iman Anum and Khatri, Swara Punit and Nawaz, Shayan and Makhdoom, Qurat Ulain and Wazir, Shahab and Patel, Kirtan and Ghaly, Mohamd},
  journal={Journal of the Egyptian National Cancer Institute},
  volume={37},
  number={1},
  pages={20},
  year={2025},
  publisher={Springer}
}

@article{tudela2024complete,
  title={A complete benchmark for polyp detection, segmentation and classification in colonoscopy images},
  author={Tudela, Yael and Maj{\'o}, Mireia and de la Fuente, Neil and Galdran, Adrian and Krenzer, Adrian and Puppe, Frank and Yamlahi, Amine and Tran, Thuy Nuong and Matuszewski, Bogdan J and Fitzgerald, Kerr and others},
  journal={Frontiers in oncology},
  volume={14},
  pages={1417862},
  year={2024}
}

@article{jha2021comprehensive,
  title={A comprehensive study on colorectal polyp segmentation with ResUNet++, conditional random field and test-time augmentation},
  author={Jha, Debesh and Smedsrud, Pia H and Johansen, Dag and De Lange, Thomas and Johansen, H{\aa}vard D and Halvorsen, P{\aa}l and Riegler, Michael A},
  journal={IEEE journal of biomedical and health informatics},
  volume={25},
  number={6},
  pages={2029--2040},
  year={2021},
  publisher={IEEE}
}

@article{wang2025dcatnet,
  title={DCATNet: polyp segmentation with deformable convolution and contextual-aware attention network},
  author={Wang, Zenan and Li, Tianshu and Liu, Ming and Jiang, Jue and Liu, Xinjuan},
  journal={BMC Medical Imaging},
  volume={25},
  number={1},
  pages={120},
  year={2025},
  publisher={Springer}
}

@article{peng2024fine,
  title={Fine-Grained Temporal Site Monitoring in EGD Streams via Visual Time-Aware Embedding and Vision-Text Asymmetric Coworking},
  author={Peng, Fang and Shi, Hongkuan and He, Shiquan and Hu, Qiang and Li, Ting and Huang, Fan and Feng, Xinxia and Liu, Mei and Liao, Jiazhi and Li, Qiang and others},
  journal={IEEE Journal of Biomedical and Health Informatics},
  year={2024},
  publisher={IEEE}
}

@article{voitov2022cortical,
  title={Cortical feedback loops bind distributed representations of working memory},
  author={Voitov, Ivan and Mrsic-Flogel, Thomas D},
  journal={Nature},
  volume={608},
  number={7922},
  pages={381--389},
  year={2022},
  publisher={Nature Publishing Group UK London}
}

@ARTICLE{intro14,
  author={Zhuang, Yuzhou and Liu, Hong and Song, Enmin and Xu, Xiangyang and Liao, Yongde and Ye, Guanchao and Hung, Chih-Cheng},
  journal={IEEE Transactions on Radiation and Plasma Medical Sciences}, 
  title={A 3-D Anatomy-Guided Self-Training Segmentation Framework for Unpaired Cross-Modality Medical Image Segmentation}, 
  year={2024},
  volume={8},
  number={1},
  pages={33-52},
 }

@article{mosch2024alarm,
  title={Alarm management in intensive care: Qualitative triangulation study},
  author={Mosch, Lina and S{\"u}mer, Meltem and Flint, Anne Rike and Feufel, Markus and Balzer, Felix and M{\"o}rike, Frauke and Poncette, Akira-Sebastian and others},
  journal={JMIR Human Factors},
  volume={11},
  number={1},
  pages={e55571},
  year={2024},
  publisher={JMIR Publications Inc., Toronto, Canada}
}

@article{jayasekara2017risk,
  title={Risk factors for metachronous colorectal cancer or polyp: A systematic review and meta-analysis},
  author={Jayasekara, Harindra and Reece, Jeanette C and Buchanan, Daniel D and Ahnen, Dennis J and Parry, Susan and Jenkins, Mark A and Win, Aung Ko},
  journal={Journal of Gastroenterology and Hepatology},
  volume={32},
  number={2},
  pages={301--326},
  year={2017},
  publisher={Wiley Online Library}
}

@article{tavanapong2022artificial,
  title={Artificial intelligence for colonoscopy: Past, present, and future},
  author={Tavanapong, Wallapak and Oh, JungHwan and Riegler, Michael A and Khaleel, Mohammed and Mittal, Bhuvan and De Groen, Piet C},
  journal={IEEE journal of biomedical and health informatics},
  volume={26},
  number={8},
  pages={3950--3965},
  year={2022},
  publisher={IEEE}
}

@article{rajpurkar2022ai,
  title={AI in health and medicine},
  author={Rajpurkar, Pranav and Chen, Emma and Banerjee, Oishi and Topol, Eric J},
  journal={Nature medicine},
  volume={28},
  number={1},
  pages={31--38},
  year={2022},
  publisher={Nature Publishing Group US New York}
}

@article{zapp2022retinal,
  title={Retinal receptive-field substructure: scaffolding for coding and computation},
  author={Zapp, S{\"o}ren J and Nitsche, Steffen and Gollisch, Tim},
  journal={Trends in Neurosciences},
  volume={45},
  number={6},
  pages={430--445},
  year={2022},
  publisher={Elsevier}
}

@inproceedings{dataset_1,
  title={Kvasir-seg: A segmented polyp dataset},
  author={Jha, Debesh and Smedsrud, Pia H and Riegler, Michael A and Halvorsen, P{\aa}l and De Lange, Thomas and Johansen, Dag and Johansen, H{\aa}vard D},
  booktitle={MultiMedia modeling: 26th international conference, MMM 2020, Daejeon, South Korea, January 5--8, 2020, proceedings, part II 26},
  pages={451--462},
  year={2020},
  organization={Springer}
}

@article{dataset_2,
  title={Towards automatic polyp detection with a polyp appearance model},
  author={Bernal, Jorge and S{\'a}nchez, Javier and Vilarino, Fernando},
  journal={Pattern Recognition},
  volume={45},
  number={9},
  pages={3166--3182},
  year={2012},
  publisher={Elsevier}
}

@article{dataset_3,
  title={WM-DOVA maps for accurate polyp highlighting in colonoscopy: Validation vs. saliency maps from physicians},
  author={Bernal, Jorge and S{\'a}nchez, F Javier and Fern{\'a}ndez-Esparrach, Gloria and Gil, Debora and Rodr{\'\i}guez, Cristina and Vilari{\~n}o, Fernando},
  journal={Computerized medical imaging and graphics},
  volume={43},
  pages={99--111},
  year={2015},
  publisher={Elsevier}
}

@ARTICLE{SegNet,
  author={Badrinarayanan, Vijay and Kendall, Alex and Cipolla, Roberto},
  journal={IEEE Transactions on Pattern Analysis and Machine Intelligence}, 
  title={SegNet: A Deep Convolutional Encoder-Decoder Architecture for Image Segmentation}, 
  year={2017},
  volume={39},
  number={12},
  pages={2481-2495},
  doi={10.1109/TPAMI.2016.2644615}}

@ARTICLE{colonsegnet,
  author={Jha, Debesh and Ali, Sharib and Tomar, Nikhil Kumar and Johansen, Håvard D. and Johansen, Dag and Rittscher, Jens and Riegler, Michael A. and Halvorsen, Pål},
  journal={IEEE Access}, 
  title={Real-Time Polyp Detection, Localization and Segmentation in Colonoscopy Using Deep Learning}, 
  year={2021},
  volume={9},
  number={},
  pages={40496-40510},
  doi={10.1109/ACCESS.2021.3063716}}

@article{liu2024devil,
  title={The devil is in the boundary: Boundary-enhanced polyp segmentation},
  author={Liu, Zhizhe and Zheng, Shuai and Sun, Xiaoyi and Zhu, Zhenfeng and Zhao, Yawei and Yang, Xuebing and Zhao, Yao},
  journal={IEEE Transactions on Circuits and Systems for Video Technology},
  volume={34},
  number={7},
  pages={5414--5423},
  year={2024},
  publisher={IEEE}
}

@InProceedings{sanderson2022fcn,
author="Sanderson, Edward
and Matuszewski, Bogdan J.",
editor="Yang, Guang and Aviles-Rivero, Angelicanand Roberts, Michael and Sch{\"o}nlieb, Carola-Bibiane",
title="FCN-Transformer Feature Fusion for Polyp Segmentation",
booktitle="Medical Image Understanding and Analysis",
year="2022",
publisher="Springer International Publishing",
address="Cham",
pages="892--907",
isbn="978-3-031-12053-4"
}

@article{yue2025boundary,
  title={Boundary-Guided Feature-Aligned Network for Colorectal Polyp Segmentation},
  author={Yue, Guanghui and Wu, Shangjie and Li, Gang and Zhao, Cheng and Hao, Yi and Zhou, Tianwei and Zhao, Baoquan},
  journal={IEEE Transactions on Circuits and Systems for Video Technology},
  year={2025},
  publisher={IEEE}
}

@article{DUCK,
  title={Using DUCK-Net for polyp image segmentation},
  author={Dumitru, Razvan-Gabriel and Peteleaza, Darius and Craciun, Catalin},
  journal={Scientific reports},
  volume={13},
  number={1},
  pages={9803},
  year={2023},
  publisher={Nature Publishing Group UK London}
}

@ARTICLE{sgu,
  author={Lei, Tao and Sun, Rui and Du, Xiaogang and Fu, Huazhu and Zhang, Changqing and Nandi, Asoke K.},
  journal={IEEE Journal of Biomedical and Health Informatics}, 
  title={SGU-Net: Shape-Guided Ultralight Network for Abdominal Image Segmentation}, 
  year={2023},
  volume={27},
  number={3},
  pages={1431-1442},
  doi={10.1109/JBHI.2023.3238183}}

@inproceedings{dilated,
  title={Dilatedsegnet: A deep dilated segmentation network for polyp segmentation},
  author={Tomar, Nikhil Kumar and Jha, Debesh and Bagci, Ulas},
  booktitle={International conference on multimedia modeling},
  pages={334--344},
  year={2023},
  organization={Springer}
}

@article{AGCNet,
  title={AGCNet: a Precise adaptive global context network for real-time colonoscopy},
  author={Shi, Liantao and Li, Zhengguo and Li, Jianyang and Wang, Yufeng and Wang, Hongyu and Guo, Yubao},
  journal={IEEE Access},
  volume={11},
  pages={59002--59015},
  year={2023},
  publisher={IEEE}
}

@article{ali2024assessing,
  title={Assessing generalisability of deep learning-based polyp detection and segmentation methods through a computer vision challenge},
  author={Ali, Sharib and Ghatwary, Noha and Jha, Debesh and Isik-Polat, Ece and Polat, Gorkem and Yang, Chen and Li, Wuyang and Galdran, Adrian and Ballester, Miguel-{\'A}ngel Gonz{\'a}lez and Thambawita, Vajira and others},
  journal={Scientific Reports},
  volume={14},
  number={1},
  pages={2032},
  year={2024},
  publisher={Nature Publishing Group UK London}
}

@INPROCEEDINGS{cmunext,
  author={Tang, Fenghe and Ding, Jianrui and Quan, Quan and Wang, Lingtao and Ning, Chunping and Zhou, S. Kevin},
  booktitle={2024 IEEE International Symposium on Biomedical Imaging (ISBI)}, 
  title={CMUNEXT: An Efficient Medical Image Segmentation Network Based on Large Kernel and Skip Fusion}, 
  year={2024},
  volume={},
  number={},
  pages={1-5},
  doi={10.1109/ISBI56570.2024.10635609}}

@ARTICLE{intro27,
  author={Gu, Zaiwang and Cheng, Jun and Fu, Huazhu and Zhou, Kang and Hao, Huaying and Zhao, Yitian and Zhang, Tianyang and Gao, Shenghua and Liu, Jiang},
  journal={IEEE Transactions on Medical Imaging}, 
  title={CE-Net: Context Encoder Network for 2D Medical Image Segmentation}, 
  year={2019},
  volume={38},
  number={10},
  pages={2281-2292},
  doi={10.1109/TMI.2019.2903562}}

@InProceedings{related_work13,
author="Zhou, Zongwei
and Rahman Siddiquee, Md Mahfuzur
and Tajbakhsh, Nima
and Liang, Jianming",
title="UNet++: A Nested U-Net Architecture for Medical Image Segmentation",
booktitle="Deep Learning in Medical Image Analysis and Multimodal Learning for Clinical Decision Support",
year="2018",
publisher="Springer International Publishing",
address="Cham",
pages="3--11",
}

@article{kamara2025mdpnet,
  title={MDPNet: Multiscale Dynamic Polyp-focus Network for Enhancing Medical Image Polyp Segmentation},
  author={Kamara, Alpha Alimamy and He, Shiwen and Fofanah, Abdul Joseph and Xu, Rong and Chen, Yuehan},
  journal={IEEE Transactions on Medical Imaging},
  year={2025},
  publisher={IEEE}
}

@article{dataset_4,
  title={A benchmark for endoluminal scene segmentation of colonoscopy images},
  author={V{\'a}zquez, David and Bernal, Jorge and S{\'a}nchez, F Javier and Fern{\'a}ndez-Esparrach, Gloria and L{\'o}pez, Antonio M and Romero, Adriana and Drozdzal, Michal and Courville, Aaron},
  journal={Journal of healthcare engineering},
  volume={2017},
  number={1},
  pages={4037190},
  year={2017},
  publisher={Wiley Online Library}
}

\begin{IEEEbiography}[{\includegraphics[width=1in,height=1.25in,clip,keepaspectratio]{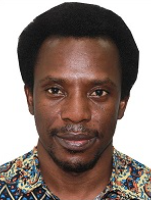}}]{Abdul Joseph Fofanah}
(Member, IEEE) earned an associate degree in mathematics from Milton Margai Technical University in 2008, a B.Sc. (Hons.) degree and M.Sc. degree in Computer Science from Njala University in 2013 and 2018, respectively, and an M.Eng. degree in Software Engineering from Nankai University in 2020. Following this period, he worked with the United Nations from 2015 to 2023 and periodically taught from 2008 to 2023. He is currently pursuing a Ph.D. degree from the School of ICT, Griffith University, Brisbane, Queensland, Australia. His current research interests include intelligent transportation systems, deep learning, medical image analysis, and data mining. 
\end{IEEEbiography}

\begin{IEEEbiography}[{\includegraphics[width=1in,height=1.25in,clip,keepaspectratio]{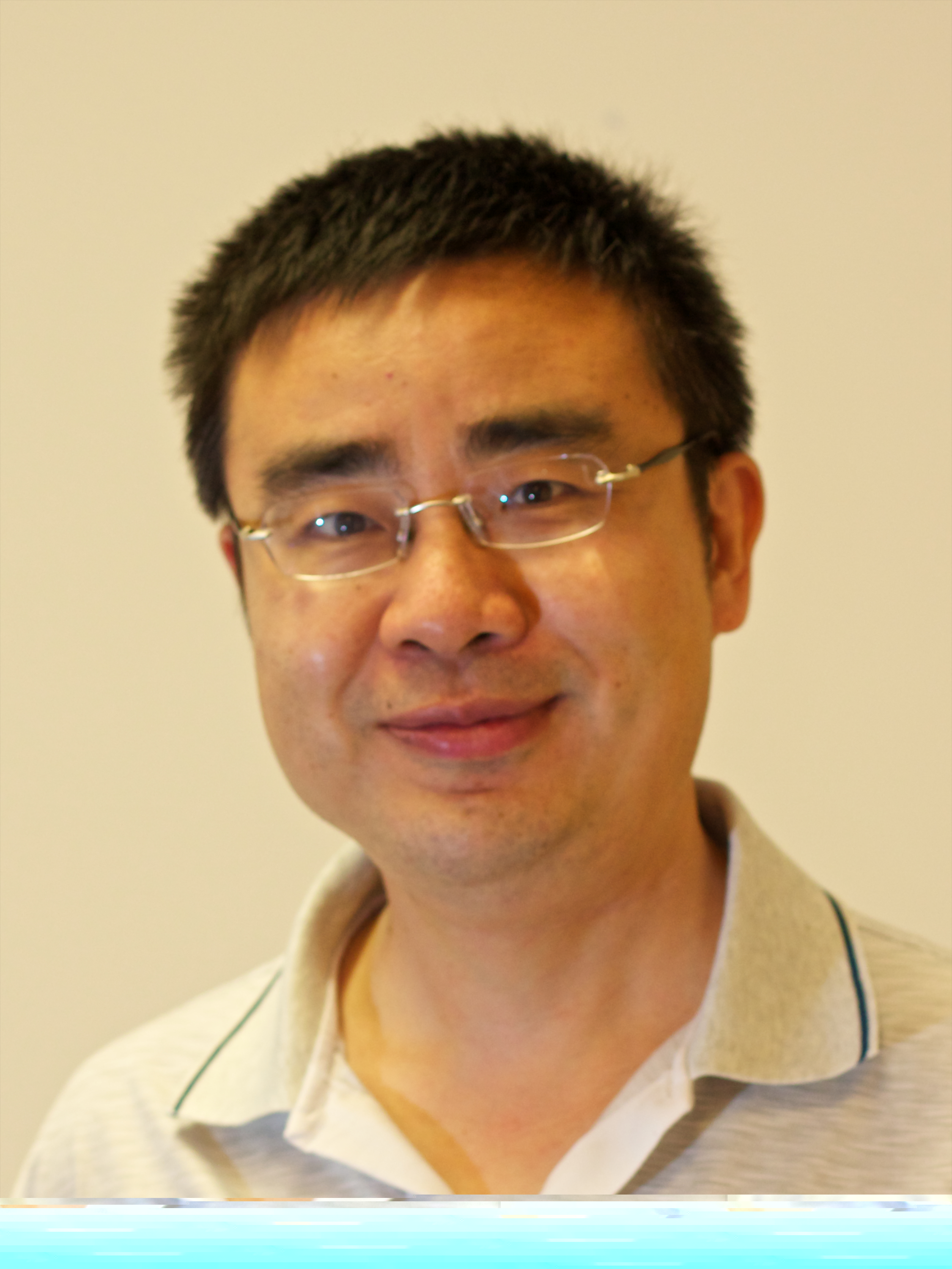}}]{Lian Wen (Larry)}
(Member, IEEE) is currently a Lecturer at the School of ICT at Griffith University. He earned a Bachelor’s degree in Mathematics from Peking University in 1987, followed by a Master’s degree in Electronic Engineering from the Chinese Academy of Space Technology in 1991. Subsequently, he worked as a Software Engineer and Project Manager across various IT companies before completing his Ph.D in Software Engineering at Griffith University in 2007. Larry’s research interests span four key areas: Software Engineering: Focused on Behaviour Engineering, Requirements Engineering, and Software Processes, Complex Systems and Scale-Free Networks, Logic Programming: With a particular emphasis on Answer Set Programming, Generative AI and Machine Learning.

\end{IEEEbiography}

\begin{IEEEbiography}[{\includegraphics[width=1in,height=1.25in,clip,keepaspectratio]{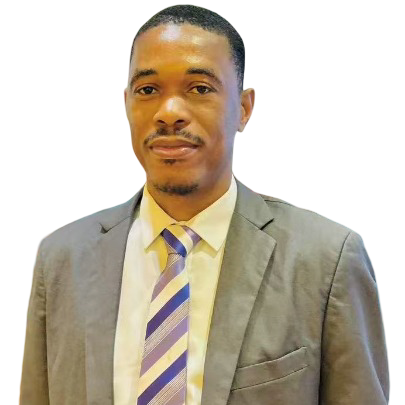}}]{Alpha Alimamy Kamara} earned a B.Eng. (Hons.) degree in Electrical and Electronics Engineering from the University of Sierra Leone in 2015 and an M.Eng. degree in Software Engineering from Nankai University in 2021. He worked as a Radio Network Planning and Optimisation Engineer at Africell from 2016 to 2019 and later as a Principal Lecturer at Limkokwing University of Creative Technology, Sierra Leone, from 2021 to 2023. His current research interests include deep learning, medical image analysis, computer vision, and multimodal image fusion.
\end{IEEEbiography}

\begin{IEEEbiography}[{\includegraphics[width=1in,height=1.25in,clip,keepaspectratio]{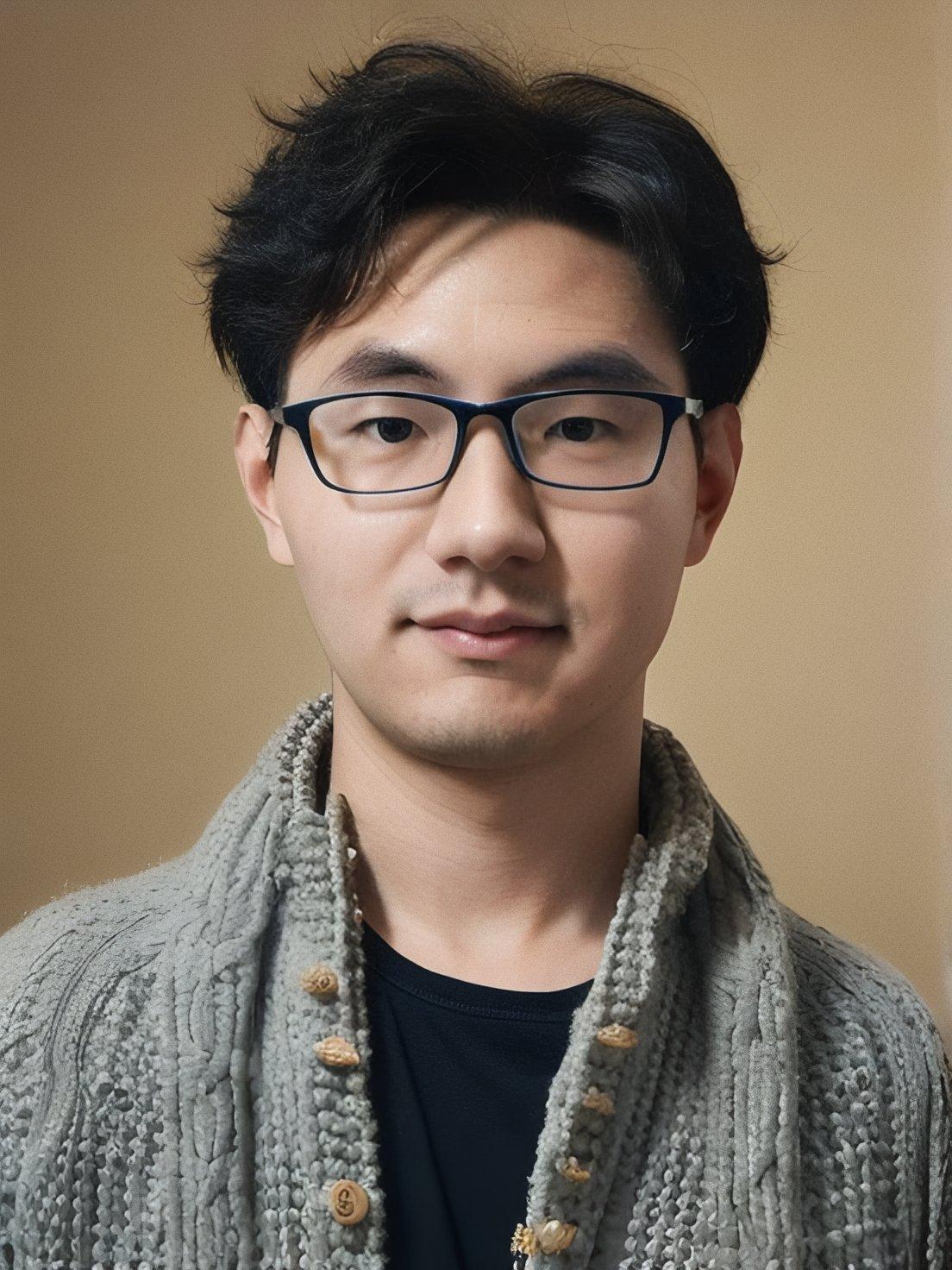}}]{Zhongyi Zhang} received a B.Eng. degree in Automation from Shandong University in 2018 and an M.Res. degree in Biomedical Data Science from Imperial College London in 2019. He worked as a Research Assistant at Harvard Medical School, Boston, from 2020 to 2021. Since 2023, he has been pursuing a Ph.D. degree with the School of Information and Communication Technology, Griffith University, Brisbane, QLD, Australia. His current research interests include deep learning, computer vision, and medical image analysis.
\end{IEEEbiography}

\begin{IEEEbiography}[{\includegraphics[width=1in,height=1.25in,clip,keepaspectratio]{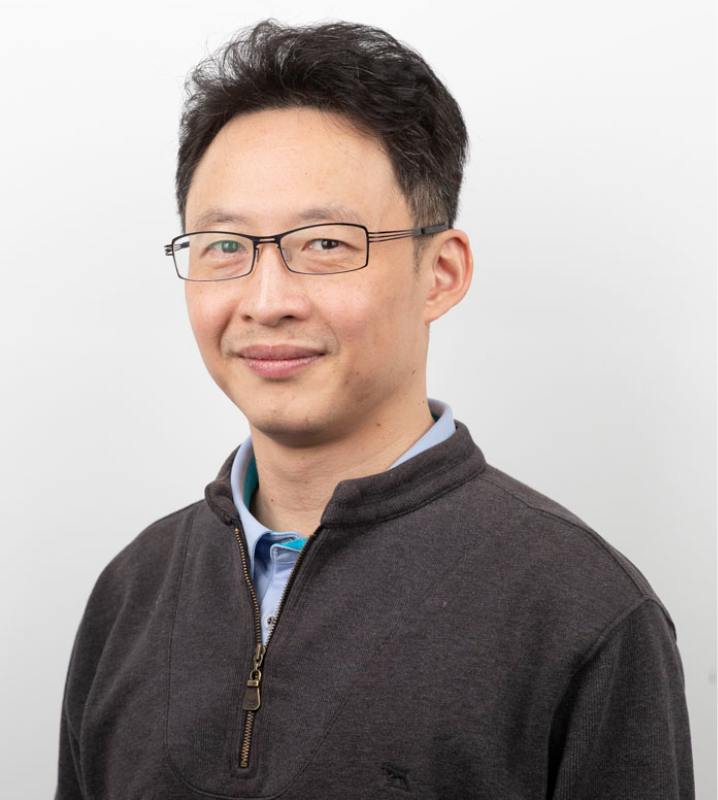}}]{David Chen}
 (Member, IEEE) obtained his Bachelor with first class Honours in 1995 and PhD in 2002 in Information Technology from Griffith University. He worked in the IT industry as a Technology Research Officer and a Software Engineer before returning to academia. He is currently a senior lecturer and serving as the Program Director for Bachelor of Information Technology in the School of Information and Communication Technology, Griffith University, Australia. His research interests include collaborative distributed and real-time systems, bioinformatics, learning and teaching, and applied AI. 
\end{IEEEbiography}

\begin{IEEEbiography}[{\includegraphics[width=1in,height=1.25in,clip,keepaspectratio]{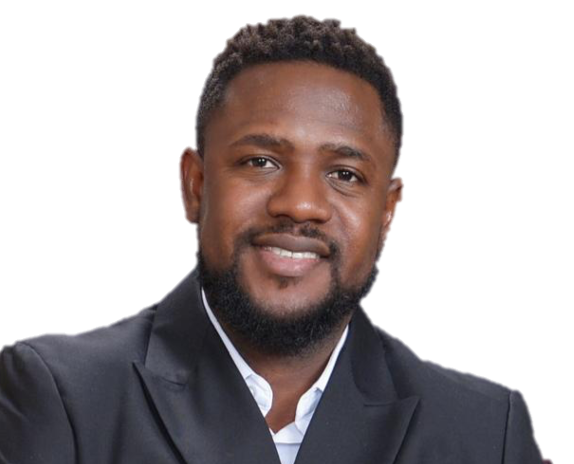}}]{Albert Partick Sankoh}
 (Senior Member, IEEE) earned a B.Sc. degree in Information Systems from the University of Sierra Leone in 2011 and an M.S. degree in Artificial Intelligence from Northeastern University, Boston, in 2024. He served as Head of IT at Access Bank Sierra Leone and founded Yestech Solutions in 2019, where he is CEO. In the U.S., he worked as an AI intern and later full-time AI Solutions Developer at Victory Human Services. He is currently Co-Founder and CTO of EZ-Tech Solutions LLC, leading research and development of their flagship product, HomewiseAI. His professional interests include AI applications in healthcare, Fintech, and Real Estate Intelligence.
\end{IEEEbiography}




\vfill


\end{document}